# MagRobot: An Open Simulator for Magnetically Navigated Robots

Heng Wang[1], *Member, IEEE*, Haoyu Song[1], Jiatao Zheng[1], Yuxiang Han[1], Kunli Wang[1]

[1]Shien-Ming Wu School of Intelligent Engineering, South China University of Technology, Guangzhou, China 511442

*Abstract*—Magnetic navigation systems, including magnetic tracking systems and magnetic actuation systems, have shown great potential for occlusion-free localization and remote control of intracorporeal medical devices and robots in minimally invasive medicine, such as capsule endoscopy and cardiovascular intervention. However, the design of magnetically navigated robots remains heavily reliant on experimental prototyping, which is time-consuming and costly. Furthermore, there is a lack of a consistent experimental environment to compare and benchmark the hardware and algorithms across different magnetic navigation systems. To address these challenges, we propose the first universal open-source simulation platform to facilitate research, design and benchmarking of magnetically navigated robots. Our simulator features an intuitive graphical user interface that enables the user to efficiently design, visualize, and analyze magnetic navigation systems for both rigid and soft robots. The proposed simulator is versatile, which can simulate both magnetic actuation and magnetic tracking tasks in diverse medical applications that involve deformable anatomies. The proposed simulator provides an open development environment, where the user can load third-party anatomical models and customize both hardware and algorithms of magnetic navigation systems. The fidelity of the simulator is validated using both phantom and *ex vivo* experiments of magnetic navigation of a continuum robot and a capsule robot with diverse magnetic actuation setups. Three use cases of the simulator, i.e., bronchoscopy, endovascular intervention, and gastrointestinal endoscopy, are implemented to demonstrate the functionality of the simulator. It is shown that the configuration and algorithms of magnetic navigation systems can be flexibly designed and optimized for better performance using the simulator.

*Index Terms*—magnetic navigation systems, simulation, magnetic robots, magnetic actuation, magnetic tracking

## I. Introduction

Magnetic field of various magnetic sources (e.g., permanent magnets and electromagnetic coils) can be engineered to transmit location information or force/torque wirelessly over a large distance. Since magnetic field can permeate most non-ferromagnetic materials including the human body, magnetic pose tracking systems [1-2] and magnetic actuation systems [3-6] have been increasingly developed for occlusion-free localization and remote control of intracorporeal medical robots in minimally invasive medicine. Magnetic actuation systems can exert necessary force and torque on miniature robots with an onboard magnetic object for precise locomotion and dexterous manipulation. For example, a capsule endoscope can be magnetically controlled to inspect the gastrointestinal tracts [7-12]. A flexible catheter can be magnetically steered to navigate through a complex vascular network in endovascular interventions [13-15]. Magnetic tracking systems can provide accurate pose information of medical robots for image registration, surgical guidance, and feedback control. Since both magnetic actuation systems and magnetic tracking systems help guide medical robots to navigate through tortuous and complex intra-body environments, they are generally referred to as "magnetic navigation systems". In this paper, the medical robots that use one or both of magnetic tracking systems and magnetic actuation systems are called magnetically navigated robots.

In magnetic actuation systems, a magnetic source (i.e., magnetic actuator) is used to apply the necessary magnetic field and gradient at the location of the robot in order to generate desired magnetic force and torque on the robot. Typical magnetic actuators include permanent magnets (PM) [16-22], electromagnets (EM) [23-28], and Helmholtz-Maxwell (H-M) coils [29-33]. For permanent-magnet-based actuators, the magnetic field and gradient are controlled by adjusting the pose of the permanent magnet using a mechanism such as a serial manipulator. For electromagnet-based actuators and Helmholtz-Maxwell coils, the magnetic field and gradient are mainly controlled by adjusting the current of the coils. A magnetic object needs to be embedded in the robot to interact with the external field for magnetic actuation. For rigid capsule robots embedded with a permanent magnet, both translation and rotation of the robot can be controlled by magnetic actuation. For continuum robots such as flexible catheters and needles, the purpose of magnetic actuation is normally to steer the robot tip and guide the robot through the lumen fluently without collisions. Extensive research has been conducted on the design of magnetic actuators [18-19] [24-26], magnetic robots [11] [12] [22] [33], and magnetic control algorithms [8] [9] [15] [17] [27] to improve the dexterity, accuracy, range, and efficiency of magnetic actuation.

Magnetic tracking systems have been widely used in applications of medical robot navigation, where various medical devices are localized including puncture needles [34], catheters [35], endoscopes [36], and surgical tools [37]. In magnetic tracking systems, the magnetic field of a magnetic object is modeled and measured for pose estimation. Existing magnetic tracking systems can be classified as permanent-magnet-based systems [38-41] and electromagnet-based systems [42-49]. In PM based systems, a permanent magnet is attached to the target object (e.g., a medical robot or instrument) and an array of sensors nearby are used to measure its magnetic field for pose estimation. The PM based system has the advantage of requiring only a passive magnet attached to the target object. However, the PM based system can only achieve five degrees-of-freedom (5-DoF) pose tracking due to the rotational symmetry of magnetic field about the axis of magnetic moment. In addition, PM based pose estimation suffers from surrounding ferromagnetic disturbances due to the use of DC magnetic field. To achieve full 6-DoF pose estimation and prevent ferromagnetic disturbances, EM based systems are developed, where a magnetic sensor is attached to the moving target to measure the AC magnetic field of multiple electromagnetic coils for pose determination. Most commercial magnetic pose tracking systems are EM based systems, e.g., NDI Aurora.

Although researchers have developed a variety of magnetic navigation systems to guide different medical robots for diverse medical procedures, there exist several major challenges that hinders the development and application of magnetically navigated medical robots. First, the design and development of magnetic robots highly relies on experiments, which require a large amount of time and cost to prototype, test, and improve the system through multiple iterations. Second, since the experimental environments are not consistent for different magnetic robot systems, it is difficult to fairly compare the performance of different systems. There is a lack of standard environment to benchmark the performance of magnetic navigation systems. Third, the learning curve for the highly sophisticated magnetic navigation systems is very steep for medical professionals with limited knowledge of magnetic systems and robotics, which significantly prevents their clinical use. To solve the challenges above, computer simulation emerges as a promising low-cost way of designing, benchmarking, and learning magnetic navigation systems for medical robots.

Multiple simulators, e.g., ROS Gazebo [50] and CoppeliaSim (formerly known as V-REP) [51], have been developed to facilitate the design and prototyping of general robotic systems including robot manipulators and mobile robots. In medical robotics, there are a few commercial simulators such as SimNow for da Vinci robots from the Intuitive Surgical company. The Surgical Science company also developed medical simulators for various procedures including sonography, endovascular intervention, endoscopic surgery, and laparoscopic surgery. These commercial simulators are closed-source and designed to support only a specific procedure with a fixed physical setting. The main purpose of these simulators is training health professionals rather than assisting with the design of medical robot systems. There are also some open-source medical simulators for research purposes. Based on V-REP, Fontanelli et al. [52] developed a simulator for the da Vinci Research Kit (dVRK [53]) tele-robotic platform, a research platform consisting of hardware components from retired da Vinci robots. Munawar et al. proposed an Asynchronous Multi-Body Framework (AMBF) for real-time simulation of dynamics, control and machine learning of the dVRK system [54] [55] [56]. Connecting the physics environment of the da Vinci robot with the learning interface of OpenAI Gym, Richter et al. [57] proposed an open-source simulator, dVRL, for reinforcement learning in surgical robotics. Xu et al. [58] developed an open-source embodied AI simulator, SurRoL, which is focused on reinforcement learning and compatible with dVRK. The open-source simulators above are all designed for simulation of rigid-link tele-robotic systems such as da Vinci robots. While AMBF supports simulation of soft and deformable anatomies, the other simulators above can only simulate manipulating rigid-body objects.

Simulators have also been developed for magnetic capsule robots and magnetic continuum robots. Abu-Kheil et al. [59] developed a simulator for magnetic endoscopic capsules. Dreyfus et al. [60] proposed a simulation framework for magnetic continuum robots based on SOFA (Simulation Open Framework Architecture [61]), a well-known open-source simulation framework for multi-model interactive physical simulation that can simulate both soft and rigid body dynamics. Although these simulators are pioneering works of simulation of magnetically navigated robots, they are still special-purpose ones for simulation of a specific type of magnetic robot using a specific magnetic actuation setup. Therefore, they are not open and flexible for the user to design and customize their own magnetic navigation systems. In addition, they cannot simulate magnetic tracking tasks. Furthermore, these simulators assume a rigid anatomy and do not consider tissue deformation in interaction with the robot.

In this work, we propose a universal open-source simulation platform for magnetically navigated robots, MagRobot (https://github.com/MagRobotics/MagRobot). Here are the major contributions of this simulator.

1. The proposed simulator is the first universal simulation platform for magnetically navigated robots to the best of our knowledge, where the user can efficiently design, visualize, and analyze magnetic navigation systems for both rigid untethered robots and flexible continuum robots through a standard workflow of pre-processing, computation and post-processing.
2. The proposed simulator is versatile because it supports both magnetic actuation and magnetic pose tracking tasks in diverse medical applications including capsule endoscopy, bronchoscopy, cardiovascular intervention, endoscopic surgery, etc.
3. The proposed simulator is realistic and computationally efficient. Enabled by SOFA, real-time simulation and visualization of magnetic navigation tasks are achieved, and tissue and robot deformation can be simulated with high fidelity as the robot interacts with the soft anatomy.
4. The proposed simulator provides an open development environment, where the user can customize magnetic pose estimation and motion control algorithms through an open interface, import third-party or customized geometric models of anatomies, and export simulation data for cross-platform use.
5. In addition to design of magnetic navigation systems, the proposed simulator can also be used to benchmark both the hardware configuration and algorithms of magnetic navigation systems in a standard simulation environment. The simulator can also be connected with various input and display devices for training health professionals on magnetic robotic systems.

This paper is organized as follows: Section II presents the general system architecture of the simulation platform. Two major computation modules for magnetic navigation tasks, i.e., magnetic actuation module and magnetic localization module, are introduced in Section III and Section IV, respectively. The physical models of robots and magnetic sources as well as motion control and pose estimation algorithms are also demonstrated. In Section V, the fidelity of the proposed simulator is validated using experiments. Section VI gives three typical use cases of the simulator, which show how it can assist in the design of magnetic navigation systems for medical robots. The paper is concluded in Section VII.

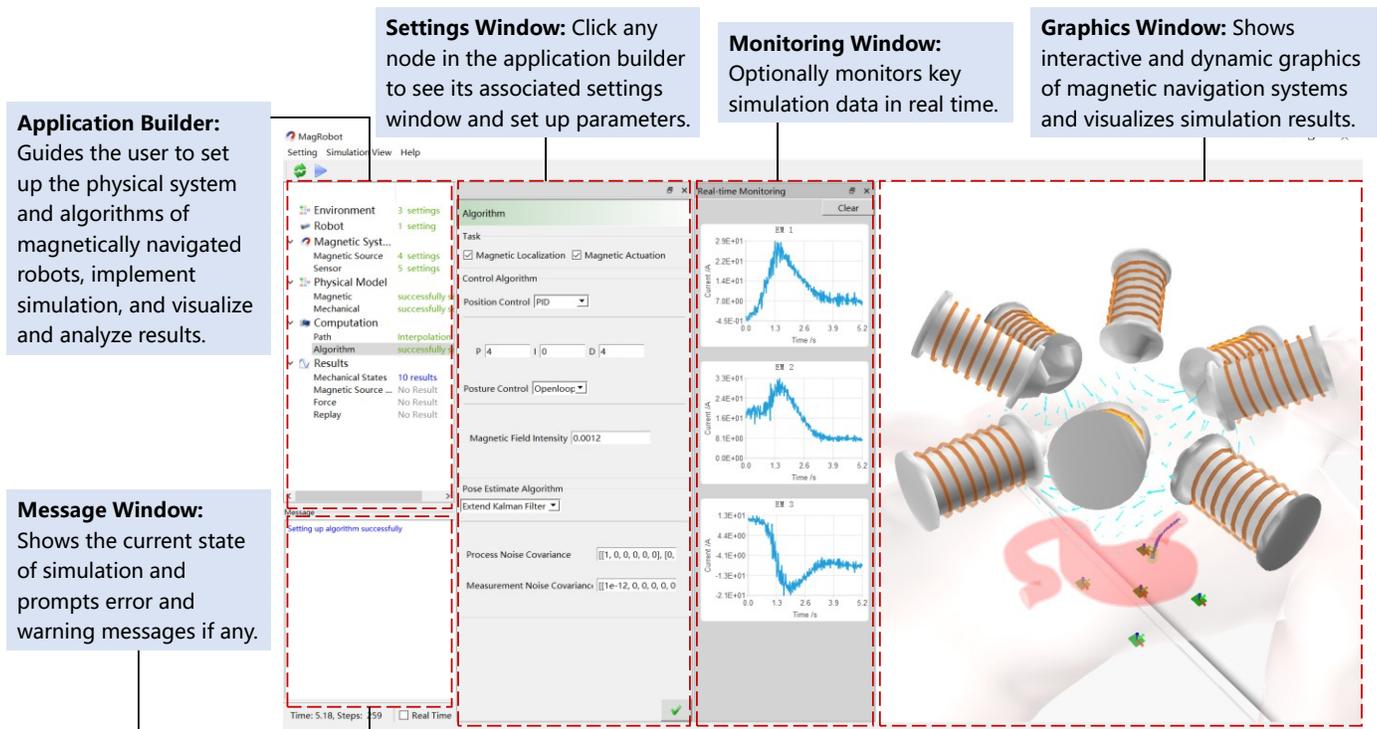

Fig. 1. The graphical user interface (GUI) of MagRobot simulator.

This article was conducted in accordance with the Declaration of Helsinki and approved by the Ethics Committee of South China University of Technology (No. 2025102).

## II. SYSTEM ARCHITECTURE

The major purpose of the proposed simulator is to facilitate the design and development of magnetic navigation systems for miniature medical robots. The simulator is built with physical models of magnetic systems and robot dynamics to accurately compute the behavior of magnetic navigation of various small-scale medical robots. In the simulator, the user can design and test both hardware configuration and algorithms of the magnetic navigation system. As shown in Fig. 1, the simulator has a graphic user interface (GUI) through which the user has the flexibility to set up the simulation environment and parameters and view the simulation results of magnetic navigation. The simulator takes the three-stage workflow as in the general computer aided engineering (CAE) software, i.e., preprocessing, computation, and postprocessing. The architecture of the simulator is organized according to the three-stage workflow, as shown in Fig. 2.

(1) *Preprocessing Stage*: The simulation environment and the robot model are created with their geometric and physical properties specified. Magnetic devices, such as electromagnetic coils and magnetic sensors, can be defined and placed in proper locations or on the robot for magnetic actuation and localization. Two major tasks of magnetic navigation, i.e., magnetic actuation and magnetic localization, are supported by the simulator and can be implemented independently or simultaneously in a closed-loop system.

(2) *Computation Stage*: The user needs to generate the trajectory of the robot first using the trajectory generator module. Then the magnetic actuation module and the magnetic tracking module can be employed to simulate the actuation and localization of the robot, respectively. The interaction (collision and friction) between the robot and the soft tissue is also considered in the simulator and the resulting deformation of soft bodies (including the soft robot and the soft tissue) is computed in real time using the SOFA framework [61] combined with the BeamAdapter plugin [62]. The magnetic navigation of the robot can be visualized, and key states and parameters of motion can be monitored in real time.

(3) *Postprocessing Stage*: The performance of the simulated magnetic navigation system is visualized and evaluated. Errors of motion control and localization are counted and plotted. 3D trajectories, magnetic field and force field can be displayed to the user for analysis and optimization of the magnetic navigation system.

Next, the detailed function of each stage of the simulator is explained.

*A. Preprocessing Stage*

In the preprocessing stage, the simulation project is created and configured. The configuration of a simulation project of magnetic navigation includes creating and defining the simulation environment, the robot, the magnetic system, and the task. The environment defines the region of robot motion and can interact with the robot. Although any rigid or soft object can be a valid environment in the simulator, the environment normally refers to a 3D anatomical model of a human lumen where magnetic medical robots navigate and work. In the simulator, three typical lumens are supported including the gastrointestinal tract, the cardiovascular system, and the respiratory tracts. The meshed 3D geometric models of these anatomies are provided in the STL and MSH format and can be easily imported into the simulator. The simulator also accepts customized or third-party models as the environment. The environment can be configured as fixed or not fixed. For a non-fixed rigid body environment, inertial parameters need to be specified. For a locally fixed or unfixed soft body environment, the mechanical properties like Young's modulus and Poisson's ratio need to be specified in addition to inertial parameters. After the environment is imported and configured, the type of the small medical robot needs to be specified. Two types of magnetic medical robots are supported in the simulator, i.e., the rigid capsule robots and the flexible continuum robots. The capsule robots are widely used for endoscopic examination and biopsy of the gastrointestinal tract and the continuum robots are widely used as flexible manipulators for transluminal and endoluminal interventions. The capsule robot is treated as a rigid body, whose inertial parameters need to be specified. The continuum robot is regarded as a soft rod, whose mechanical properties like Young's modulus need to be specified in addition to inertial parameters. In addition, multi-joint robot manipulators can also be created in the simulator for moving the magnetic sources of the magnetic actuation system. Once the robot is specified and created, the next step is to create magnetic devices including magnetic sources and magnetic sensors. Electromagnetic coils with and without a ferrous core and permanent magnets can be created with their geometric dimension, pose and magnetic moment (intensity) defined. Depending on the configuration of the magnetic actuator, the magnetic source can be stationary or placed on a moving mechanism (e.g., a manipulator). Typical configurations of the magnetic actuator, e.g., moving permanent magnets, multiple stationary electromagnetic coils, and Maxwell-Helmholtz coils, can be easily designed and simulated. Magnetic sensors can be created to measure the magnetic field from magnetic sources for localization with their number, location, and noise levels specified. Finally, the user needs to specify the simulation task as magnetic actuation or magnetic localization. It is noted that magnetic actuation and localization can also be simultaneously simulated in a closed loop.

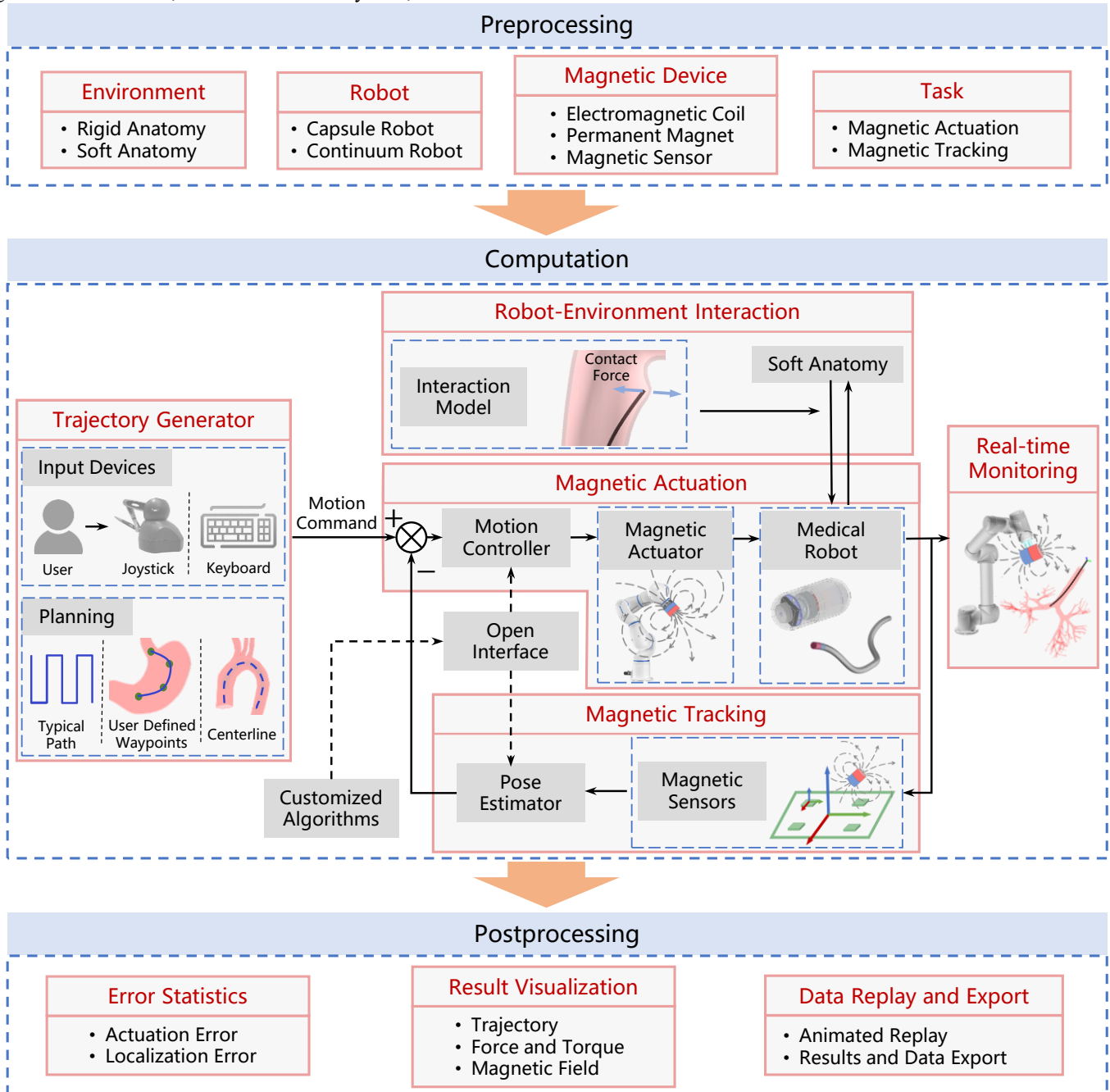

Fig. 2. Architecture and workflow of MagRobot simulator.

*B. Computation Stage*

In the computation stage, magnetic actuation and localization of the magnetic robot are simulated using the built-in physical models of the magnetic navigation system and robots. The interaction between the robot and the deformable environment (soft lumen) is computed using the SOFA framework. Default motion controller, pose estimator, and trajectory planner are provided to enable closed-loop magnetic actuation, magnetic localization, and trajectory generation. Users are also allowed to develop and test their customized algorithms of motion control, pose estimation and path planning through open interfaces. There are five modules in the computation engine: (1) trajectory generator module, (2) magnetic actuation module, (3) magnetic tracking module, (4) robot-environment interaction module, and (5) visual monitoring module.

(1) *Trajectory Generator*: This module is used to generate the trajectory of the robot to navigate through the lumen. The generated trajectory provides the motion command for the motion control of magnetic navigation. Three ways of trajectory generation are supported in the simulator. First, the trajectory of the robot can be dynamically controlled by the user through an input device such as a keyboard or a joystick (e.g., haptic device) in a teleoperation mode. The user can also generate the full trajectory ahead of simulation. A second way of trajectory generation is to input the position of waypoints or click to choose waypoints on the GUI and then interpolate between the waypoints using B-spline curves [63]. In the third way, the trajectory can be automatically generated using the centerline-based path planning algorithm [14], where the centerline of the lumen is extracted as the path of robot navigation to prevent collision with the lumen. The extraction of centerlines is implemented using the VMTK toolkit [64]. In the simulator, the constant-velocity trajectory is generated by default according to the planned path.

(2) *Magnetic Actuation*: This module is used to simulate the magnetic actuation task. Given the motion command from the trajectory generator and the feedback of robot pose, the motion controller computes the control input of the magnetic actuator, e.g., the current supply to the electromagnetic coils or the pose of the manipulator that holds the permanent magnet. Then, the resulting magnetic field and magnetic force/torque applied to the magnetic robot can be computed according to the physical model of the magnetic actuator. Finally, the robot motion is computed subject to the magnetic force/torque and the interaction force with the environment using dynamic models of a rigid body (capsule robot) or a soft body (continuum robot).

(3) *Magnetic Tracking*: This module is used to simulate the magnetic tracking task. Magnetic sensors can be placed near or on the robot to measure the magnetic field of the magnetic source for pose estimation. The estimated pose can provide feedback for motion control in the actuation module. The tracking module can also be decoupled from the actuation module and implemented alone. Both electromagnet-based and permanent-magnet-based pose tracking can be simulated. The user can either place the magnetic source on the moving robot and use stationary sensors to track its motion or place the magnetic sensor on the moving robot to measure the magnetic field of one or more stationary magnets.

(4) *Robot-environment Interaction*: This module addresses the contact interaction between the robot and the environment using the built-in collision solvers of SOFA. Both rigid and soft environments (anatomies) can be simulated in the proposed platform. The collision between the robot and the environment is detected first and the contact points are determined at the interface. The contacts are treated as constraints to the dynamics of the rigid or soft bodies. The contact forces including both the normal collision force and the friction can be solved using a Lagrange multipliers approach [61]. Once the contact forces are known, the corrective motion of the medical robot and the soft anatomy is solved to obtain their deformation.

(5) *Visual Monitoring*: This module enables the user to monitor the simulated scene of the robot and the soft lumen in real time. The motion and deformation can be visualized in a graphic viewer with the potential contact forces illustrated by arrows. The time variation of kinematic states (i.e., position, velocity, and acceleration) and forces can be plotted in real time. The visual monitoring module can provide the immediate indication of the magnetic navigation performance so the user can efficiently diagnose problems and optimize the system early before the postprocessing stage.

*C. Postprocessing Stage*

In the postprocessing stage, simulation results can be visualized, analyzed, and exported for cross-platform processing. The postprocessed results provide useful feedback to the user for further optimization of the magnetic navigation system. The major tasks of postprocessing are introduced below.

(1) *Error Statistics*: The error of localization and motion control can be counted and plotted in the postprocessing stage to evaluate the performance of magnetic navigation. The localization error is defined as the difference between the estimated pose and the true pose of the robot.

$$\mathbf{e}_\text{L} = [e_{\text{L},x}\ e_{\text{L},y}\ e_{\text{L},z}\ e_{\text{L},o}]^\text{T} \tag{1}$$

where $[e_{\text{L},x}\ e_{\text{L},y}\ e_{\text{L},z}]^\text{T}$ represents the position error of localization in $x$-, $y$-, and $z$-axis, and $e_{\text{L},o}$ represents the orientation error of localization. Normally, we use the axis of magnetic moment of the magnet on the robot to represent the robot orientation and thus the orientation error is the angle between the estimated direction of moment and the true direction of moment. The motion control error of magnetic actuation is defined as the difference between the true pose and the desired (planned) pose.

$$\mathbf{e}_A = [e_{A,x} \ e_{A,y} \ e_{A,z} \ e_{A,o}]^T \quad (2)$$

where $[e_{A,x} \ e_{A,y} \ e_{A,z}]^T$ represent the position error of actuation, $e_{A,o}$ represents the orientation error of actuation. Both the maximum and average error (e.g. root-mean-square error) are counted.

$$e_{\max} = \max(|e_k|)$$
$$e_{\text{avg}} = \sqrt{\frac{1}{N}\sum_{i=1}^{N} e_k^2} \quad (3)$$

where $e_k$ is the position or orientation error of the $k$-th time step.

(2) *Result Visualization*: The user can replay the animation of the dynamic simulation in the postprocessing stage. As shown in Fig. 3, the magnetic field and the force field (i.e., magnetic force, normal collision force, friction, etc.) can also be visualized using arrows overlaid on the simulation scene. The 3D trajectory of magnetic localization and actuation can be visualized using the Matplotlib library [65]. 2D plots of magnetic states, kinematic states, and forces over time can also be generated using the QChart module of the Qt framework [66].

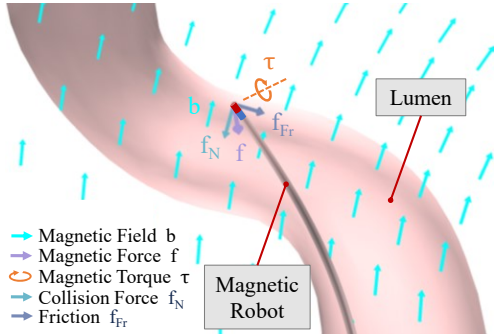

Fig. 3. Result visualization of magnetic navigation in the postprocessing stage (Using magnetic actuation of a catheter robot as an example).

(3) *Data Export*: The simulation data can be exported and stored in CSV format for further analysis and processing in other software or platforms. The visual results, e.g., the plots, images, and animations, can also be exported.

### III. MAGNETIC ACTUATION SYSTEM

In this section, the physical model of the magnetic actuator and the motion control algorithm for magnetic actuation systems are introduced.

*A. Physical Model of Magnetic Actuator*

Magnetic actuators are magnetic sources with additional mechanisms that can generate required magnetic fields and gradients for the magnetic actuation task. Typical magnetic actuators include permanent magnets, electromagnets, and Helmholtz-Maxwell coils. For permanent-magnet-based actuators, the magnetic field and gradient are controlled by adjusting the pose of the permanent magnet using a mechanism such as a serial manipulator. For electromagnet-based actuators and Helmholtz-Maxwell coils, the magnetic field and gradient are mainly controlled by adjusting the current of the coils, although they can also be moved by a mechanism for magnetic field control. The physical model of the magnetic actuator describes the relationship of the magnetic force and torque on a magnetic object (e.g., a magnetic robot) and the control input to the magnetic actuator (i.e., current and pose of magnetic sources). This model is composed of two parts: (1) The distribution of magnetic field and gradient depends on the control input to the magnetic actuator; (2) The magnetic force and torque are determined by the interaction between the local field and the magnetic robot. The former is given first for different types of actuators and the latter is given next.

(1) *Permanent-magnet Actuator*: The actuating permanent magnet is considered as a magnetic dipole, with its dipole moment denoted by $\mathbf{m}_a \in \mathbb{R}^3$ and its position denoted by $\mathbf{p}_a \in \mathbb{R}^3$. Let $\mathbf{p}_r \in \mathbb{R}^3$ be the position of a magnetic robot to be actuated. The magnetic field $\mathbf{b} \in \mathbb{R}^3$ at $\mathbf{p}_r$ can be described as:

$$\mathbf{b}(\mathbf{p}_r; \mathbf{p}_a, \mathbf{m}_a) = \frac{\mu_0}{4\pi}\left[\frac{3(\mathbf{m}_a \cdot \mathbf{r})\mathbf{r}}{\|\mathbf{r}\|^5} - \frac{\mathbf{m}_a}{\|\mathbf{r}\|^3}\right] \quad (4)$$

where $\mu_0$ is the magnetic permeability in vacuum and $\mathbf{r} = \mathbf{p}_r - \mathbf{p}_a$. The field-derivative matrix $G \in \mathbb{R}^{3\times 3}$ at $\mathbf{p}_r$ is given by

$$G\{\mathbf{r}, \mathbf{m}_a\} = \left[\frac{\partial \mathbf{b}}{\partial x} \ \frac{\partial \mathbf{b}}{\partial y} \ \frac{\partial \mathbf{b}}{\partial z}\right]$$
$$= \frac{3\mu_0}{4\pi \|\mathbf{r}\|^4}\left(\mathbf{m}_a \hat{\mathbf{r}}^T + \hat{\mathbf{r}}\mathbf{m}_a^T + (\hat{\mathbf{r}}^T \mathbf{m}_a)(I_3 - 5\hat{\mathbf{r}}\hat{\mathbf{r}}^T)\right) \quad (5)$$

In the application of magnetic actuation where no current or electric field is involved in the workspace, the following constraints on the magnetic field can be derived from Maxwell's equations.

$$\nabla \cdot \mathbf{b} = 0$$
$$\nabla \times \mathbf{b} = 0 \quad (6)$$

Therefore, there are only five independent entries in the field-derivative matrix $G$, which can be put into a field-derivative vector $\mathbf{g} \in \mathbb{R}^5$.

$$\mathbf{g}(\mathbf{p}_r; \mathbf{p}_a, \mathbf{m}_a) = \left[\frac{\partial b_x}{\partial x} \ \frac{\partial b_x}{\partial y} \ \frac{\partial b_x}{\partial z} \ \frac{\partial b_y}{\partial y} \ \frac{\partial b_y}{\partial z}\right]^T \quad (7)$$

Overall, the magnetic field and gradient of the permanent-magnet-based actuator can be put into the following form.

$$\begin{bmatrix}\mathbf{b}\\\mathbf{g}\end{bmatrix} = B_p(\mathbf{p}_r; \mathbf{p}_a, \mathbf{m}_a) \quad (8)$$

(2) *Electromagnet Actuator*: Assuming that the electromagnets are stationary, the magnetic field and gradient at $\mathbf{p}_r \in \mathbb{R}^3$ is linear with respect to the current supply to the electromagnets. Let $\mathbf{I} = [i_1 \cdots i_n]^T$ be the currents of $n$ electromagnets. The magnetic field $\mathbf{b}$ and the gradient $\mathbf{g}$ at $\mathbf{P}_r$ can be described as:

$$\begin{bmatrix}\mathbf{b}\\\mathbf{g}\end{bmatrix} = B_e(\mathbf{p}_r; \mathbf{I}) = \bar{B}(\mathbf{p}_r)\mathbf{I}$$
$$= \begin{bmatrix}\bar{\mathbf{b}}_1 & \cdots & \bar{\mathbf{b}}_j & \cdots & \bar{\mathbf{b}}_n \\ \bar{\mathbf{g}}_1 & \cdots & \bar{\mathbf{g}}_j & \cdots & \bar{\mathbf{g}}_n\end{bmatrix}\begin{bmatrix}i_1\\\vdots\\i_j\\\vdots\\i_n\end{bmatrix} \quad (9)$$

where $\bar{\mathbf{b}}_j$ and $\bar{\mathbf{g}}_j$ are the field and gradient generated by the $j$-th electromagnet as its current is set to be 1 A, respectively. When the size of the electromagnets is relatively small with respect to the workspace, the electromagnets can be regarded as dipoles. Then $\bar{\mathbf{b}}_j$ and $\bar{\mathbf{g}}_j$ takes the form of (4) and (5) with the magnetic moment given by the unit current supply.

(3) *Helmholtz-Maxwell Coils:* Helmholtz coils are a pair of circular coils of equal radius arranged coaxially, which run the same current in the same handedness. Helmholtz coils are used to generate a uniform magnetic field with no gradient in the central region. The magnetic field of Helmholtz coils is aligned with the axial direction and its magnitude is proportional to the current of the coils.

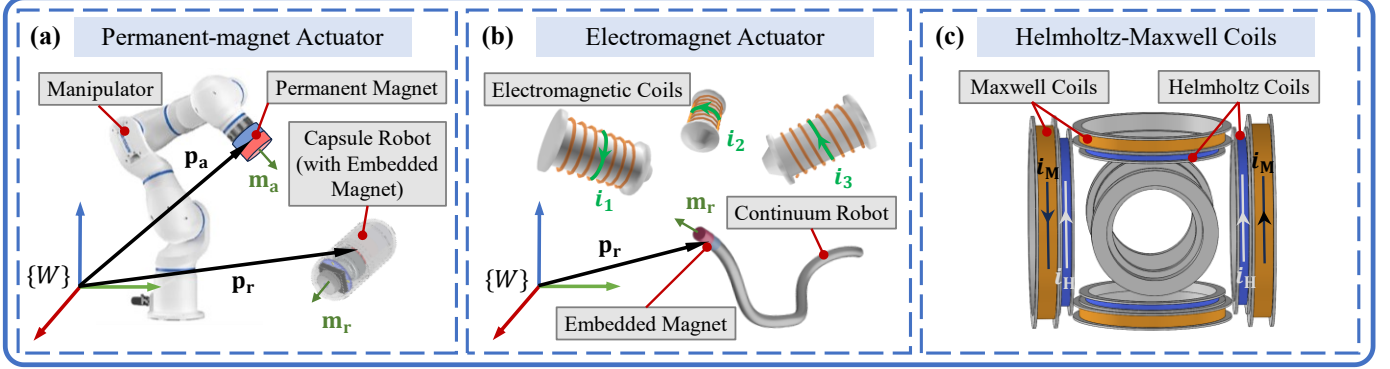

Fig. 4. Schematics of Magnetic Actuators. (a) Permanent-magnet Actuator. (b) Electromagnet Actuator. (c) Helmholtz-Maxwell Coils.

$$\|\mathbf{b}\| = (0.72\mu_0/r_H)i_H \quad (10)$$

where $r_H$ is the coil radius, which is also equal to the distance between the two parallel coils, and $i_H$ is the current of the coil. The Maxwell coils are also a pair of coaxial circular coils of equal radius, which however run the same current with the opposite handedness. In this arrangement, the magnetic field cancels to zero at the center point. The magnetic gradient of the Maxwell coils is approximately uniform in the central region, as given by

$$\frac{\partial b_z}{\partial z} = (0.64\mu_0/r_M^2)i_M \quad (11)$$

where $r_M$ is the radius and $i_M$ is the current of the Maxwell coils. The distance between two parallel coils is $\sqrt{3}r_M$. Multiple Helmholtz and Maxwell coils can be arranged in various configurations to provide magnetic fields and gradients in different directions [29]. The magnetic field and gradient of the Helmholtz-Maxwell coil system can also be described by (9) while $B(\mathbf{p}_r)$ takes the form of (10) and (11) rather than a dipole model.

Once the magnetic field and gradient are determined, the magnetic force and torque on a magnetic robot can be calculated. It is assumed that the small magnet attached to the robot is modeled as a dipole. Let the magnetic moment of the magnetic robot be $\mathbf{m}_r \in \mathbb{R}^3$. The magnetic force $\mathbf{f}$ on the robot is given by

$$\mathbf{f} = \nabla(\mathbf{b} \cdot \mathbf{m}_r) = M_G(\mathbf{m}_r)\mathbf{g} \quad (12)$$

where

$$M_G(\mathbf{m}_r) = \begin{bmatrix} m_{rx} & m_{ry} & m_{rz} & 0 & 0 \\ 0 & m_{rx} & 0 & m_{ry} & m_{rz} \\ -m_{rz} & 0 & m_{rx} & -m_{rz} & m_{ry} \end{bmatrix} \quad (13)$$

The magnetic torque $\boldsymbol{\tau}$ on the robot is given by

$$\boldsymbol{\tau} = \mathbf{m}_r \times \mathbf{b} = S(\mathbf{m}_r)\mathbf{b} \quad (14)$$

where

$$S(\mathbf{m}_r) = \begin{bmatrix} 0 & -m_{rz} & m_{ry} \\ m_{rz} & 0 & -m_{rx} \\ -m_{ry} & m_{rx} & 0 \end{bmatrix} \quad (15)$$

Then the full physical model of the magnetic actuator is obtained as follows.

$$\begin{aligned} \begin{bmatrix} \boldsymbol{\tau} \\ \mathbf{f} \end{bmatrix} &= M_{\tau,f}\{\mathbf{m}_r\} \begin{bmatrix} \mathbf{b} \\ \mathbf{g} \end{bmatrix} \\ &= M_{\tau,f}\{\mathbf{m}_r\} B(\mathbf{p}_r; \mathbf{w}) \\ &= F_{\tau,f}(\mathbf{m}_r, \mathbf{p}_r; \mathbf{w}) \end{aligned} \quad (16)$$

where $\mathbf{w}$ is the control parameter of the magnetic actuator (i.e., $\mathbf{p}_a$, $\mathbf{m}_a$ and $\mathbf{I}$) and

$$M_{\tau,f}\{\mathbf{m}_r\} = \begin{bmatrix} S\{\mathbf{m}_r\} & 0_{3\times 5} \\ 0_{3\times 3} & M_G\{\mathbf{m}_r\} \end{bmatrix} \quad (17)$$

$$B(\mathbf{p}_r; \mathbf{w}) = B_p(\mathbf{p}_r; \mathbf{p}_a, \mathbf{m}_a) \text{ or } B_e(\mathbf{p}_r; \mathbf{I}) \quad (18)$$

depending on the type of magnetic actuator.

*B. Physical Model of Robot*

The physical model of robots describes the motion of capsule and continuum robots subject to external forces, including magnetic forces (torques), gravity and constraint forces from the environment.

(1) *Capsule Robot:* Capsule robots are modeled as a rigid body with 6-DoF motion. The dynamics of translation is given by the Newton's second law of motion.

$$\mathbf{F} = m_r \mathbf{a}_r \quad (19)$$

where $\mathbf{F}$ is the total force exerted on the robot, $m_r$ is the mass of the robot, and $\mathbf{a}_r$ is the acceleration of the robot. Once the acceleration $\mathbf{a}_r$ is solved, the position and velocity of the robot can be obtained by integrating the following ordinary differential equation.

$$\begin{bmatrix} \dot{\mathbf{p}}_r \\ \dot{\mathbf{v}}_r \end{bmatrix} = \begin{bmatrix} 0_{3\times 3} & I_3 \\ 0_{3\times 3} & 0_{3\times 3} \end{bmatrix} \begin{bmatrix} \mathbf{p}_r \\ \mathbf{v}_r \end{bmatrix} + \begin{bmatrix} 0_{3\times 3} \\ \mathbf{a}_r \end{bmatrix} \quad (20)$$

The rotation dynamics of the capsule robot is given by the Euler equation.

$$\mathbf{T} = I_C \dot{\boldsymbol{\omega}}_r + \boldsymbol{\omega}_r \times I_C \boldsymbol{\omega}_r \quad (21)$$

where $\mathbf{T}$ is the total torque acting about the center of mass of the robot, $\boldsymbol{\omega}_r$ and $\dot{\boldsymbol{\omega}}_r$ are the angular velocity and acceleration of the robot, and $I_C$ is the inertia tensor of the robot about the center of mass. Once the angular velocity is solved from equation (21), the orientation of the robot can be computed using the following dynamic equation.

$${}^n_r\dot{\mathbf{R}} = S(\boldsymbol{\omega}_r){}^n_r\mathbf{R} \quad (22)$$

where ${}^n_r\mathbf{R}$ denotes the orientation of the robot represented in the inertial frame.

(2) *Continuum Robot:* In the simulator, the motion and deformation of the continuum robots subject to external

forces is computed using the *BeamAdapter* plugin in the SOFA framework [62]. The continuum robots made of light materials are characterized by high tensile strength and low resistance to bending [67]. Therefore, they can be modeled as elastic rods. The one-dimensional finite element method of the continuum robot is employed based on the Kirchhoff rod theory [68], where the robot is discretized into $n$ nodes of beam elements. For the $e$-th beam element, a symmetric stiffness matrix $K_e$, which depends on elasticity, inertia, and geometry of the robot, is used to relate the 6-DoF displacements (three translations and three rotations) of two end nodes and the forces and torques applied to them, as given by

$$\mathbf{F}_e = K_e \mathbf{U}_e \quad (23)$$

where $\mathbf{U}_e$ is the displacements of beam nodes corresponding to the nodal forces $\mathbf{F}_e$. Then the equilibrium equation of the whole continuum robot can be established by assembling the elementary equations for each beam element [62].

$$\mathbf{F} = K\mathbf{U} \quad (24)$$

where $\mathbf{U}$ is the displacements of the robot corresponding to the external forces $\mathbf{F}$. $K$ is the global stiffness matrix that sums the contribution of all beam elements. Then the deformation of the robot can be computed by inverting the equation (24).

*C. Motion Control Algorithm*

The motion controller provides the required control input to the magnetic actuator to drive the robot along a desired trajectory. In the default motion control algorithm of the simulator, the desired magnetic force or torque is given by a proportional-integral-derivative (PID) controller based on the desired robot pose and the feedback of the actual robot pose. Then the control input is computed by inverting the physical model of the magnetic actuator. The simulator also provides an open interface for the user to develop customized control algorithms.

(1) *Capsule Robot:* Capsule robots can be considered as a rigid body, whose orientation tends to align with the local magnetic field. Therefore, the orientation control is achieved by the open-loop control method, which sets the direction of the magnetic field $\hat{\mathbf{b}}^*$ as the desired robot orientation $\hat{\mathbf{m}}_r^*$.

$$\hat{\mathbf{b}}^* = \hat{\mathbf{m}}_r^* \quad (25)$$

The desired magnetic force $\mathbf{f}^*$ is determined using the PID feedback controller.

$$\mathbf{f}^* = k_{p,\mathbf{f}} \mathbf{e}_\mathbf{p} + k_{d,\mathbf{f}} \frac{d\mathbf{e}_\mathbf{p}}{dt} + k_{i,\mathbf{f}} \int_0^t \mathbf{e}_\mathbf{p} d\tau \quad (26)$$

where $\mathbf{e}_\mathbf{p} = \mathbf{p}_r^* - \mathbf{p}_r$ is the error of the desired robot position and the actual robot position. $k_{p,\mathbf{f}}$, $k_{i,\mathbf{f}}$ and $k_{d,\mathbf{f}}$ are the proportional, integral and derivative gains of PID controller for position control. Then the following model is inverted to find the required control input $\mathbf{w}$ to the actuator.

$$\begin{bmatrix}\hat{\mathbf{b}}^*\\\mathbf{f}^*\end{bmatrix} = M_{\mathbf{b},\mathbf{f}}\{\mathbf{m}_r\}\begin{bmatrix}\hat{\mathbf{b}}\\\mathbf{g}\end{bmatrix}$$
$$= M_{\mathbf{b},\mathbf{f}}\{\mathbf{m}_r\}B'(\mathbf{p}_r;\mathbf{w}) \quad (27)$$
$$= F_{\mathbf{b},\mathbf{f}}(\mathbf{m}_r,\mathbf{p}_r;\mathbf{w})$$

where

$$M_{\mathbf{b},\mathbf{f}}\{\mathbf{m}_c\} = \begin{bmatrix} I_3 & 0_{3\times 5} \\ 0_{3\times 3} & M_G\{\mathbf{m}_c\} \end{bmatrix} \quad (28)$$

It is noted that $B'(\mathbf{p}_r;\mathbf{w})$ is different from $B(\mathbf{p}_r;\mathbf{w})$ because the magnetic field is of unit magnitude. Since model (27) is nonlinear in terms of control input $\mathbf{w}$, linearization is conducted to facilitate solution of the inverse problem.

$$\begin{bmatrix}\delta\hat{\mathbf{b}}^*\\\delta\mathbf{f}^*\end{bmatrix} = J_{F_{\mathbf{b},\mathbf{f}}}(\mathbf{m}_r,\mathbf{p}_r;\mathbf{w})\delta\mathbf{w} \quad (29)$$

where $J_{F_{\mathbf{b},\mathbf{f}}}(\mathbf{m}_r,\mathbf{p}_r;\mathbf{w})$ is the Jacobian matrix of the function $F_{\mathbf{b},\mathbf{f}}(\mathbf{m}_r,\mathbf{p}_r;\mathbf{w})$. The pseudoinverse solution of the incremental control input is given by

$$\delta\mathbf{w} = J_{F_{\mathbf{b},\mathbf{f}}}(\mathbf{m}_r,\mathbf{p}_r;\mathbf{w})^\dagger \begin{bmatrix}\delta\hat{\mathbf{b}}^*\\\delta\mathbf{f}^*\end{bmatrix} \quad (30)$$

where "†" denotes the pseudoinverse of a matrix.

(2) *Continuum Robot:* In the teleoperation mode, the advancement and bidirectional bending of the robot can be controlled by the user using an input device such as a keyboard. In this human-in-the-loop method, the user can adjust the direction of magnetic field until it can bend the robot into the desired orientation. In the autonomous navigation mode, the advancement length in each step is determined as the distance between the neighboring waypoints of the trajectory. The desired magnetic steering torque $\boldsymbol{\tau}^*$ is determined using the PID feedback controller.

$$\boldsymbol{\tau}^* = k_{p,\boldsymbol{\tau}}\mathbf{e}_\mathbf{m} + k_{d,\boldsymbol{\tau}}\frac{d\mathbf{e}_\mathbf{m}}{dt} + k_{i,\boldsymbol{\tau}}\int_0^t \mathbf{e}_\mathbf{m} d\tau \quad (31)$$

where $\|\mathbf{e}_\mathbf{m}\| = \angle \mathbf{m}_r^*$, $\mathbf{m}_r$ is the angle between the desired robot orientation $\mathbf{m}_r^*$ and the actual robot orientation $\mathbf{m}_r$, and $\hat{\mathbf{e}}_\mathbf{m} = \widehat{\mathbf{m}_r \times \mathbf{m}_r^*}$. $k_{p,\boldsymbol{\tau}}$, $k_{i,\boldsymbol{\tau}}$ and $k_{d,\boldsymbol{\tau}}$ are the proportional, integral and derivative gains of the orientation controller. Denote (14) as

$$\boldsymbol{\tau} = S(\mathbf{m}_r)\mathbf{b} = F_{\boldsymbol{\tau}}(\mathbf{m}_r,\mathbf{p}_r;\mathbf{w}) \quad (32)$$

Then model (32) can be inverted to find the required control input $\mathbf{w}$ to the actuator. If a permanent-magnet actuator is used, the magnetic field is nonlinear with respect to control parameters (pose of the actuating magnet) and the pseudoinverse solution of the incremental pose of the actuating magnet is given by

$$\begin{bmatrix}\delta\mathbf{p}_a\\\delta\mathbf{m}_a\end{bmatrix} = J_{F_{\boldsymbol{\tau}}}(\mathbf{m}_r,\mathbf{p}_r;\mathbf{p}_a,\mathbf{m}_a)^\dagger \delta\boldsymbol{\tau}^* \quad (33)$$

If an electromagnet-based actuator is used, the magnetic field is linear with control parameters (currents of electromagnets) and the current can be directly solved as follows.

$$\mathbf{I} = \left(S(\mathbf{m}_r)\bar{\mathbf{b}}\right)^\dagger \boldsymbol{\tau}^* \quad (34)$$

In the default control algorithm, the magnetic force is not controlled in both teleoperation and autonomous mode.

IV. MAGNETIC TRACKING SYSTEM

In this section, the magnetic measurement models and the pose estimation algorithms for magnetic tracking systems are introduced.

*A. Measurement Model of Magnetic Tracking Systems*

The major components of magnetic tracking systems include magnetic sources (e.g., electromagnets or permanent magnets) and magnetic sensors. In applications of robot localization, either a magnetic sensor or a magnetic source is placed on the robot, while an array of stationary magnetic sources or sensors are placed nearby. The sensor measures the magnetic field signal that varies with the robot pose. The magnetic measurement model describes the relationship between the magnetic measurements and the robot pose. Once the measurement model is established, an estimation algorithm is implemented to recover the robot pose from the magnetic field measurements according to the model. In the simulator, two typical magnetic tracking systems are supported, including the one with a magnet on the robot and the one with a magnetic sensor on the robot. The magnetic measurement models for these two types of tracking systems are shown below.

(1) *Magnetic Source on Robot*: In the first configuration of magnetic tracking systems, multiple magnetic sensors are used to measure the magnetic field of the magnetic source on the robot, as shown in Fig. 5(a). The magnetic source can either be a permanent magnet or an electromagnet, but a permanent magnet is preferred since it is wireless. The magnetic source is assumed as a dipole, which generates a rotationally symmetric magnetic field around the magnet. Therefore, the orientation around the magnetic moment axis cannot be determined from magnetic measurements and only 5-DoF pose estimation is feasible for this configuration. Let the magnetic moment of the magnetic source for localization be $\mathbf{m}_L \in \mathbb{R}^3$. Then the orientation of the magnet can be denoted by the unit vector $\hat{\mathbf{m}}_L$. Let the position of the magnet be $\mathbf{p}_L \in \mathbb{R}^3$. Let the position and orientation of the $i$-th sensor be $\mathbf{p}_{s,i} \in \mathbb{R}^3$ and $_w^s\mathbf{R}_i \in SO(3)$, $i = 1,2,...,n$. The subscript "w" represents the world frame and the superscript "s" represents the sensor frame. Then the magnetic measurement by the $i$-th sensor is given by

$$\mathbf{b}_{s,i}(\mathbf{p}_L, \hat{\mathbf{m}}_L) = \frac{\mu_0}{4\pi} {_w^s}\mathbf{R}_i \left[ \frac{3(\mathbf{m}_L \cdot \mathbf{r}_{L,i})\mathbf{r}_{L,i}}{\|\mathbf{r}_{L,i}\|^5} - \frac{\mathbf{m}_L}{\|\mathbf{r}_{L,i}\|^3} \right] \quad (35)$$

where $\mathbf{r}_{L,i} = \mathbf{p}_{s,i} - \mathbf{p}_L$. It is noted that $_w^s\mathbf{R}_i$ and $\mathbf{p}_{s,i}$ are constants and the states to be estimated are $\mathbf{p}_L$ and $\hat{\mathbf{m}}_L$. Then the measurement model can be obtained by (36).

$$\mathbf{B}_m(\mathbf{p}_L, \hat{\mathbf{m}}_L) = \begin{bmatrix} \mathbf{b}_{s,1}(\mathbf{p}_L, \hat{\mathbf{m}}_L) \\ \mathbf{b}_{s,2}(\mathbf{p}_L, \hat{\mathbf{m}}_L) \\ \cdots \\ \mathbf{b}_{s,n}(\mathbf{p}_L, \hat{\mathbf{m}}_L) \end{bmatrix} \quad (36)$$

(2) *Magnetic Sensor on Robot*: In the second configuration, a tri-axis magnetic sensor is put on the robot to measure the magnetic fields from multiple stationary electromagnetic coils for pose tracking, as shown in Fig. 5(b). Different electromagnets are excited sequentially (time division multiplexing) or simultaneously in different frequencies (frequency division multiplexing) to distinguish the magnetic signals. In this configuration, the 6-DoF pose of the sensor (robot) can be estimated. Let the orientation of the sensor be represented by the unit quaternion $\mathbf{q}_s \in SO(3)$.

$$\mathbf{q}_s = [q_0, q_1, q_2, q_3]^T \quad (37)$$

The rotation matrix $_w^s\mathbf{R}$ is related to $\mathbf{q}_s$ by

$$_w^s\mathbf{R}(\mathbf{q}_s)$$
$$= \begin{bmatrix} 1 - 2q_2^2 - 2q_3^2 & 2q_1q_2 + 2q_3q_0 & 2q_1q_3 - 2q_2q_0 \\ 2q_1q_2 - 2q_3q_0 & 1 - 2q_2^2 - 2q_3^2 & 2q_2q_3 + 2q_1q_0 \\ 2q_1q_3 + 2q_2q_0 & 2q_2q_3 - 2q_1q_0 & 1 - 2q_2^2 - 2q_3^2 \end{bmatrix} \quad (38)$$

Let the magnetic moment and position of the $j$-th electromagnet is $\mathbf{m}_{L,j}$ and $\mathbf{p}_{L,j}$, $j = 1,2,...,m$. Then the magnetic field of the $j$-th electromagnet measured by the sensor is given by

$$\mathbf{b}_{s,j}(\mathbf{p}_s, \mathbf{q}_s) = \frac{\mu_0}{4\pi} {_w^s}\mathbf{R}(\mathbf{q}_s) \left[ \frac{3(\mathbf{m}_{L,j} \cdot \mathbf{r}_{L,j})\mathbf{r}_{L,j}}{\|\mathbf{r}_{L,j}\|^5} - \frac{\mathbf{m}_{L,j}}{\|\mathbf{r}_{L,j}\|^3} \right] \quad (39)$$

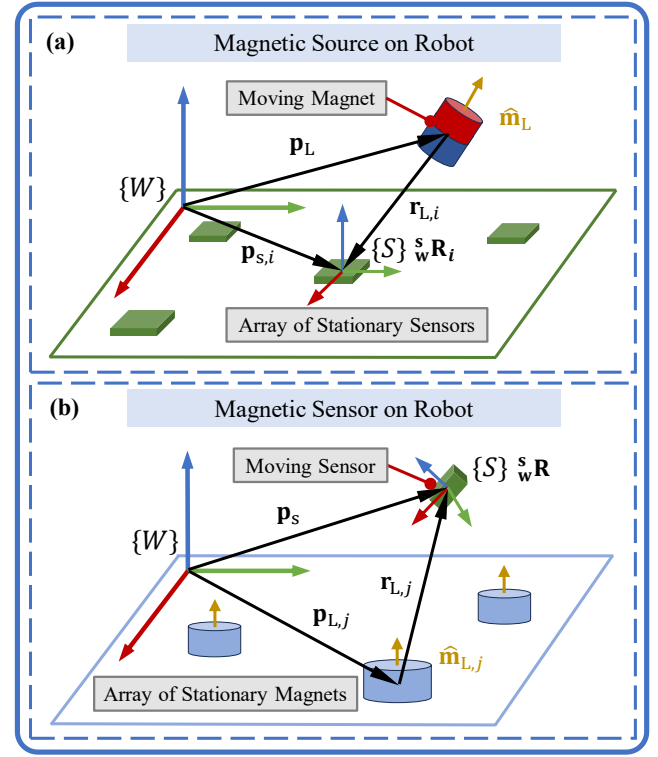

Fig. 5. (a) Schematic of the magnetic localization system using a magnet on the robot and stationary magnetic sensors. (b) Schematic of the magnetic localization system using a magnetic sensor on the robot and stationary magnetic sources.

where $\mathbf{r}_{L,j} = \mathbf{p}_s - \mathbf{p}_{L,j}$. It is noted that $\mathbf{m}_{L,j}$ and $\mathbf{p}_{L,j}$ are constants and the states to be estimated are $\mathbf{p}_s$ and $\mathbf{q}_s$. Then the measurement model is given by

$$\mathbf{B}_s(\mathbf{p}_s, \mathbf{q}_s) = \begin{bmatrix} \mathbf{b}_{s,1}(\mathbf{p}_s, \mathbf{q}_s) \\ \mathbf{b}_{s,2}(\mathbf{p}_s, \mathbf{q}_s) \\ \cdots \\ \mathbf{b}_{s,m}(\mathbf{p}_s, \mathbf{q}_s) \end{bmatrix} \quad (40)$$

*B. Pose Estimation Algorithms*

The pose estimator recovers the robot pose information from the magnetic measurements using a nonlinear estimation algorithm. There are two types of pose estimation algorithms, the Bayes filter-based algorithms and the optimization-based algorithms. In the simulator, extended Kalman filter (EKF) is used as the default algorithm of the Bayes filter type and Levenberg-Marquardt (LM) algorithm [43] [44] is used as the default algorithm of the optimization-based type. The simulator also provides an open interface for the user to develop customized estimation algorithms. For simplicity, only the EKF algorithm is shown below. The EKF algorithm has been widely used in magnetic pose estimation for its simplicity of implementation and high computational efficiency [49]. In addition, the EKF algorithm can fuse the magnetic measurements with a kinematic or dynamic model of the robot to increase the tracking accuracy and suppress the sensor noise.

For the tracking system with a magnet on the robot, the state vector to be estimated is

$$\mathbf{x} = [\mathbf{p}_L, \mathbf{v}_L, \hat{\mathbf{m}}_L]^T \quad (41)$$

where $\mathbf{v}_L$ is the linear velocity of the magnet. For the tacking system with a sensor on the robot, the state vector to be estimated is

$$\mathbf{x} = [\mathbf{p}_s, \mathbf{v}_s, \mathbf{q}_s]^T \quad (42)$$

where $\mathbf{v}_s$ is the linear velocity of the sensor. In the default EKF algorithm of the simulator, a trivial constant-velocity kinematic model (42) is used to propagate the states. If the kinematic or dynamic model of the robot is known, it can be fused with magnetic measurements instead.

$$\begin{aligned} \mathbf{x}_k &= \mathbf{A}\mathbf{x}_{k-1} + \mathbf{G}_k \mathbf{w}_{k-1} \\ \mathbf{w}_k &\sim \mathcal{N}(\mathbf{0}, \mathbf{Q}) \end{aligned} \quad (43)$$

where

$$\mathbf{A} = \begin{bmatrix} \mathbf{I} & \mathbf{I}dt & \mathbf{0} \\ \mathbf{0} & \mathbf{I} & \mathbf{0} \\ \mathbf{0} & \mathbf{0} & \mathbf{I} \end{bmatrix} \quad (44)$$

The process noise $\mathbf{w}_k$ is modeled as a zero mean Gaussian white noise consisting of the unmodelled linear acceleration with variances $\sigma_{a_x}^2, \sigma_{a_y}^2, \sigma_{a_z}^2$ and the unmodelled rotation velocity with variances $\sigma_{\omega_x}^2, \sigma_{\omega_y}^2, \sigma_{\omega_z}^2$. $\mathcal{N}(\boldsymbol{\mu}, \boldsymbol{\Sigma})$ denotes a Gaussian distribution with mean $\boldsymbol{\mu}$ and covariance $\boldsymbol{\Sigma}$. The covariance of $\mathbf{w}_k$ is given by

$$\mathbf{Q} = \text{diag}(\sigma_{a_x}^2, \sigma_{a_y}^2, \sigma_{a_z}^2, \sigma_{\omega_x}^2, \sigma_{\omega_y}^2, \sigma_{\omega_z}^2) \quad (45)$$

For the localization system with a magnet on the robot, $\mathbf{G}_k$ is given by

$$\mathbf{G}_k = \begin{bmatrix} \frac{1}{2}\mathbf{I}_3 dt^2 & \mathbf{0}_{3\times 3} \\ \mathbf{I}_3 dt & \mathbf{0}_{3\times 3} \\ \mathbf{0}_{3\times 3} & \frac{1}{2}\mathbf{C}_m(\hat{\mathbf{m}}_{L,k-1})dt \end{bmatrix} \quad (46)$$

where

$$\mathbf{C}_m(\hat{\mathbf{m}}_L) = \begin{bmatrix} 0 & m_3 & -m_2 \\ -m_3 & 0 & m_1 \\ m_2 & -m_1 & 0 \end{bmatrix} \quad (47)$$

For the localization system with a sensor on the robot, $\mathbf{G}_k$ is given by

$$\mathbf{G}_k = \begin{bmatrix} \frac{1}{2}\mathbf{I}_3 dt^2 & \mathbf{0}_{3\times 3} \\ \mathbf{I}_3 dt & \mathbf{0}_{3\times 3} \\ \mathbf{0}_{4\times 3} & \frac{1}{2}\mathbf{C}_s(\mathbf{q}_{s,k-1})dt \end{bmatrix} \quad (48)$$

where

$$\mathbf{C}_s(\mathbf{q}_s) = \begin{bmatrix} -q_1 & -q_2 & -q_3 \\ q_0 & -q_3 & q_2 \\ q_3 & q_0 & -q_1 \\ -q_2 & q_1 & q_0 \end{bmatrix} \quad (49)$$

As given by (36) and (40), the magnetic measurement model can be denoted by

$$\mathbf{h}(\mathbf{x}) = \mathbf{B}_m\{\mathbf{p}_L, \hat{\mathbf{m}}_L\} \text{ or } \mathbf{B}_s\{\mathbf{p}_s, \mathbf{q}_s\} \quad (50)$$

Considering the sensor noise, the magnetic measurement $\mathbf{z}_{s,k}$ is given by

$$\begin{aligned} \mathbf{z}_{s,k} &= \mathbf{h}(\mathbf{x}_k) + \mathbf{u}_k \\ \mathbf{u}_k &\sim \mathcal{N}(\mathbf{0}, \mathbf{U}) \end{aligned} \quad (51)$$

where

$$\mathbf{U} = \text{diag}\left(\sigma_{b_{x,1}}^2, \sigma_{b_{y,1}}^2, \sigma_{b_{z,1}}^2, \dots, \sigma_{b_{x,n}}^2, \sigma_{b_{y,n}}^2, \sigma_{b_{z,n}}^2\right) \quad (52)$$

The sensor noise of the $i$-th sensor $\mathbf{u}_{i,k}$ is also modeled as a zero-mean Gaussian white noise with variance $\sigma_{b_{x,i}}^2, \sigma_{b_{y,i}}^2, \sigma_{b_{z,i}}^2$. Once the kinematic equation (43) and the measurement equation (51) are constructed, a standard EKF algorithm can be implemented to estimate the robot pose, as given by Algorithm 1.

---

**Algorithm 1:** Extend Kalman Filter

**Input:** Previous pose $\mathbf{x}_{k-1}^+$, previous covariance $\mathbf{P}_{k-1}^+$, the measurement vector $\mathbf{z}_{s,k}$, process noise covariance $\mathbf{Q}$ and observation noise covariance $\mathbf{R}$;

**Output:** Updated pose $\mathbf{x}_k^+$ and updated covariance $\mathbf{P}_k^+$;

**Prediction Step:**
State Prediction: $\mathbf{x}_k^- = \mathbf{A}\mathbf{x}_{k-1}^+$;
Covariance Prediction: $\mathbf{P}_k^- = \mathbf{A}\mathbf{P}_{k-1}^+ \mathbf{A}^T + \mathbf{G}_k \mathbf{Q} \mathbf{G}_k^T$;

**Update Step:**
Jocobian Computation: $\mathbf{H}_k = \frac{\partial \mathbf{h}}{\partial \mathbf{x}}\big|_{\mathbf{x}=\mathbf{x}_{k-1}^+}$;
Kalman Gain Computation: $\mathbf{K}_k = \mathbf{P}_k^- \mathbf{H}_k^T \left(\mathbf{H}_k \mathbf{P}_k^- \mathbf{H}_k^T + \mathbf{U}\right)^{-1}$;
State Estimate Update: $\mathbf{x}_k^+ = \mathbf{x}_k^- + \mathbf{K}_k[\mathbf{z}_{s,k} - \mathbf{h}(\mathbf{x}_k^-)]$;
Covariance Estimate Update: $\mathbf{P}_k^+ = (\mathbf{I}_3 - \mathbf{K}_k \mathbf{H}_k)\mathbf{P}_k^-$;
**return** $\mathbf{x}_k^+$;

---

## V. EXPERIMENTAL VALIDATION OF SIMULATION PLATFORM

In this section, simulation results of magnetically navigated robots are compared with experimental results to validate the fidelity of the proposed simulator. In the first case, a continuum robot is steered by a manipulator-controlled permanent magnet to navigate through a vascular phantom. In the second case, a rigid capsule robot is manipulated by Helmholtz-Maxwell coils to move in a fluid environment.

### A. Continuum Robot

This task is to navigate a continuum robot through a rigid two-dimensional vascular phantom, which is implemented by simulation and experiment along the same trajectory and with the same setup parameters. The magnetic continuum robot is composed of a flexible catheter and a cylindrical permanent magnet embedded at the end, as shown in Fig. 6 (c). The geometric, mechanical and magnetic parameters of the magnetic continuum robot are listed in Table I. A permanent magnet with a magnetic moment of 17.5 A·m² is used as the actuating magnetic source, which is mounted on a 6-axis robotic manipulator (UR5, *Universal Robots Inc.*). The position and orientation of the actuating magnet are controlled by the manipulator to adjust the magnetic field in the workspace for magnetic navigation. A teleoperation control mode is adopted, where the user simultaneously delivers the continuum robot and steers its tip by setting a desired direction of magnetic field using a keyboard. The magnitude of the magnetic field is set to be 29 Gauss. The magnetic force is controlled to be zero to eliminate its negative impact on navigation, as given by (32). The motion control is implemented at a frequency of 20 Hz. The setup parameters of the magnetic navigation system for the continuum robot are given in Table II.

TABLE I: Robot Parameters in Two Validation Cases

| Validation I | Length (mm) | Outer Diameter (mm) | Inner Diameter (mm) | Young's Modulus (MPa) | Magnetic Moment (A·m²) |
|---|---|---|---|---|---|
| | 300 | 1.5 | 1.0 | 120 | 0.025 |
| Validation II | Mass (g) | Moment of Inertia (mg·m²) | Outer Diameter (mm) | Length (mm) | Magnetic moment (A·m²) |
| | 5.16 | (0.08,0.2,0.2) | 13 | 25 | 0.253 |

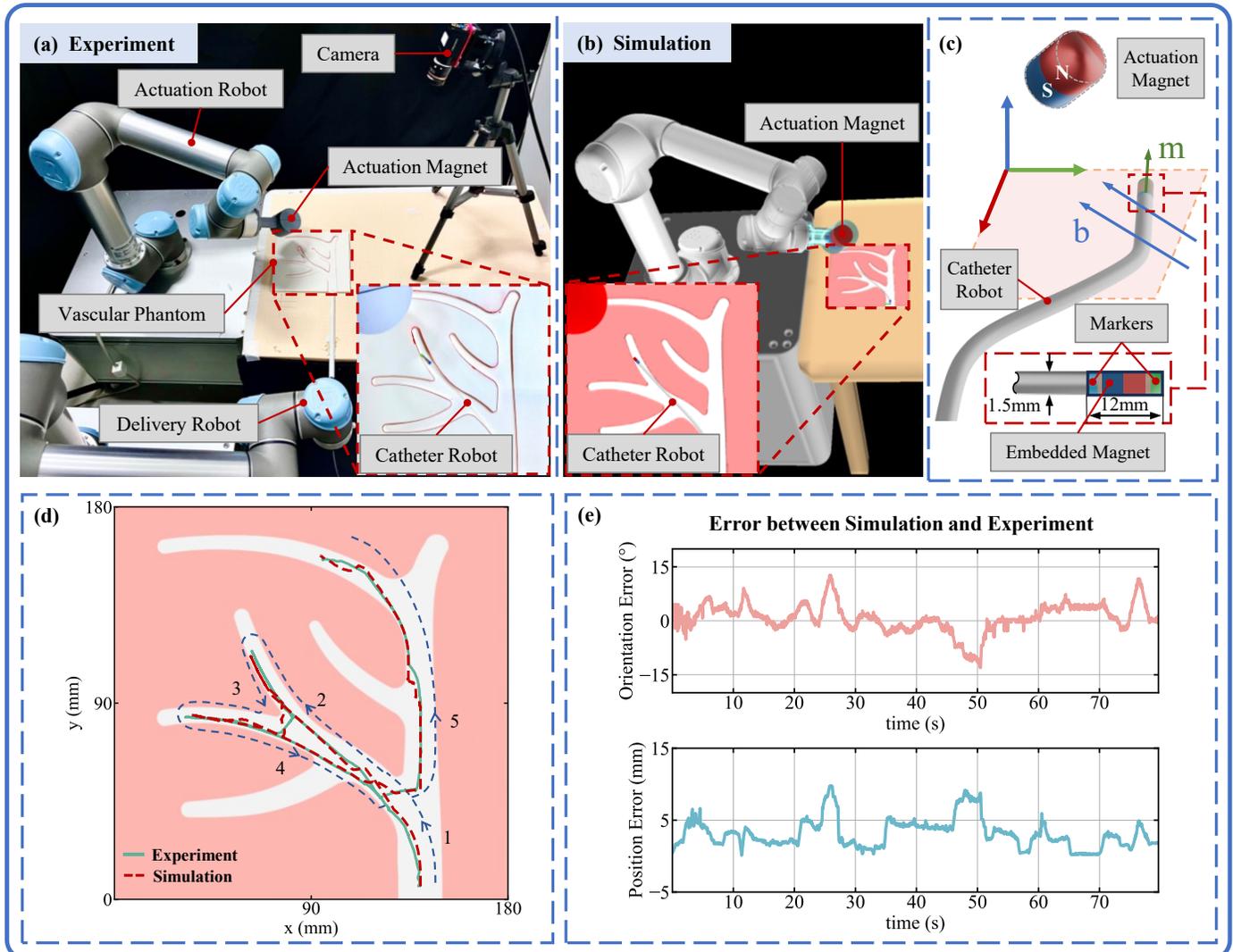

Fig. 6. (a) Experimental setup for magnetic actuation of a continuum robot. (b) Simulation scene of magnetic actuation of a continuum robot. (c) Schematic of magnetic steering of the robot tip. (d) Trajectories of the magnetic continuum robot in the experiment and simulation. (e) Trajectory errors between the experiment and the simulation.

The experiment of magnetic navigation is first implemented. Then its motion control commands given by the user are recorded and adopted in the simulation to ensure that the same trajectory is used in both simulation and experiment. As shown in Fig. 6 (a), the continuum robot is delivered by another robot manipulator (UR5, *Universal Robots Inc.*) in the experiment. The robot manipulators are communicated with and controlled by a computer (CPU: Intel i5-10210U) through Robot Operating System (ROS). An RGB camera (MV-CU004-10GM/GC, *HIKROBOT*) is used to capture the true position and orientation of the tip of the continuum robot. A green and a blue marker are attached to the robot tip for image-based pose tracking. Fig. 6 (b) shows the simulation scene, which has the same setup as in the experiment. Fig. 6 (d) shows the navigation trajectory in the simulation as compared to that in the experiment. It is shown that the continuum robot almost moves along the same trajectory in the simulator and in the experiment. As shown in Fig. 6 (e), the trajectory error between the experiment and the simulation is very small with an RMS Euclidean position error of 3.18 mm and an RMS average orientation error of 3.02°.

TABLE II
Setup Parameters of Simulation in Two Validation Cases

| Case | Environment | Robot | Magnetic Actuation Source | Trajectory | Motion Control Algorithm | Robot Localization (Experiment) | Control Frequency (Hz) | Friction |
|---|---|---|---|---|---|---|---|---|
| Validation I | Vascular phantom | Continuum robot | Manipulator-controlled PM | Teleoperation | Orientation: Human-in-the-loop | Single RGB camera | 20 | 0.1 |
| Validation II | Ex-vivo porcine stomach | Capsule robot | H-M Coils | Pre-defined square trajectory | Magnetic field: Open-loop control | Stereo high-speed cameras | 50 | 0.4 |

### B. Capsule Robot

In this task, a capsule robot is magnetically manipulated by Helmholtz-Maxwell coils to navigate in an *ex vivo* porcine stomach. The simulation and experiment are implemented along the same trajectory and with the same setup parameters. As shown in Fig. 7 (c), the rigid capsule robot is fabricated by 3D printing, with a dimension of $\varnothing 13 \times 25$ mm. A cylindrical permanent magnet with a dimension of $\varnothing 6 \times 8$ mm is embedded at the center of the capsule robot. The parameters of the capsule robot are listed in Table I. As shown in Fig. 7 (a), three pairs of Helmholtz coils and three pairs of Maxwell coils are orthogonally configured to be the source of magnetic actuation. The dimension and magnetic field/gradient per unit current of H-M coils are provided in Table III. The magnetic actuation task is conducted on the surface of an ex-vivo porcine stomach, which is located in the central workspace of the Helmholtz-Maxwell coils. A permanent magnet is embedded inside the robot for magnetic actuation. The Helmholtz coils are controlled to generate a rotating magnetic field in the central workspace in order to apply a continuing magnetic torque around the central axis of the robot and make it roll on the surface of the stomach, as shown in Fig. 7 (c). The direction of the robot can be steered by tuning the direction of rotation axis of the actuation field.

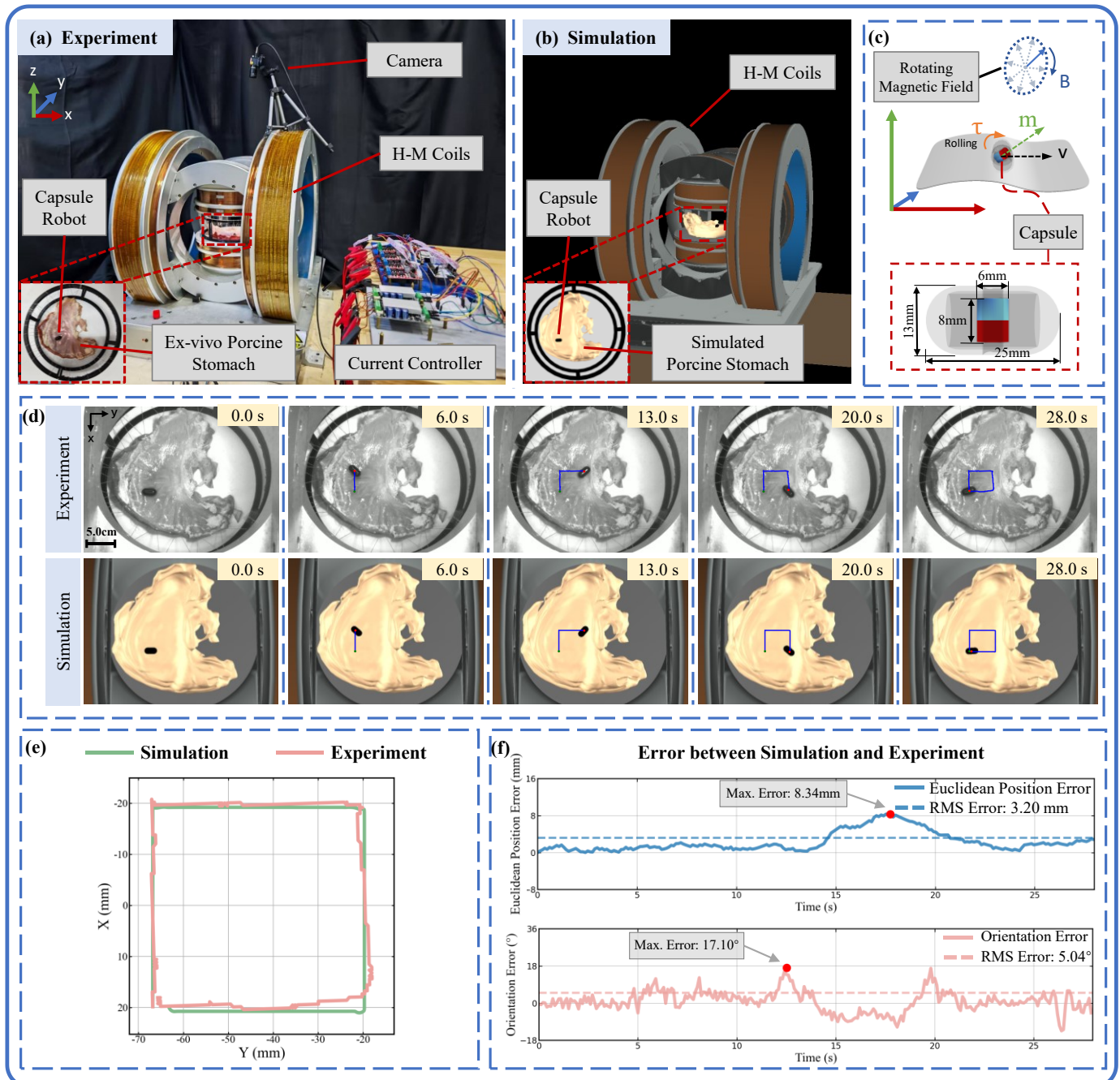

Fig. 7. (a) Experimental setup for magnetic actuation of a capsule robot. (b) Simulation scene of magnetic actuation of a capsule robot. (c) Schematic of magnetic rolling control of a capsule robot. (d) Temporal screenshots of magnetically actuated capsule robot in the simulation and experiment. (e) Trajectories of the capsule robot in the simulation and experiment. (f) Euclidean position and orientation errors between the experiment and simulation.

In this task, the capsule robot is controlled to follow a square-wave trajectory in the *xoy*-plane, with its orientation rotated by 90° at each vertex of the square path, as shown in Fig. 7 (d). It is noted that the actual trajectory is not a perfect square due to the gastric rugae and mucus on the stomach surface. One high-speed cameras (OSGO30-790UMTZ, *YVSION Inc.*) is used to localize the capsule robot. The motion control is implemented at a frequency of 50 Hz. The setup parameters of the magnetic navigation system for the capsule robot are given in Table II.

TABLE III
Parameters of H-M Coils

| Coil Type | Axis | Magnetic Field (Gauss/A) | Magnetic Field Gradient (Gauss/m/A) | Coil Radius (m) |
|---|---|---|---|---|
| Helmholtz Coils | X | 8.90 | — | 0.375 |
|  | Y | 8.80 | — | 0.265 |
|  | Z | 8.55 | — | 0.170 |
| Maxwell Coils | X | — | 18.90 | 0.560 |
|  | Y | — | 31.70 | 0.385 |
|  | Z | — | 180.0 | 0.280 |

Fig. 7 (d) shows the temporal sequence of screenshots of the magnetically actuated capsule robot in both the experiment and the simulation. It is observed that the capsule robot can follow the square trajectory well and its motion in the simulation is consistent with that in the experiment. In Fig. 7 (e), the trajectory of the capsule robot in the simulation is compared with that in the experiment. The RMS euclidean position error between the experiment and the simulation is 3.20 mm while the RMS orientation error is 5.04 degrees, as shown in Fig. 7 (f). This *ex vivo* expeiment vlidates the fidelity of the proposed simulator for the task of magnetic navigation of rigid and untethered capsule robots.

## VI. Use Cases of Simulation Platform

The simulation platform can be used by researchers and engineers to accelerate the design, development, and evaluation of magnetic navigation systems for medical robots. In this section, three case studies, i.e., bronchoscopy, endovascular intervention, and gastrointestinal endoscopy, are presented to demonstrate the functionalities of the simulation platform. These case studies involve environments with both rigid and soft human tissues, various configurations of magnetic navigation systems, and both rigid and continuum medical robots. In addition, both magnetic localization and actuation tasks are simulated with various estimation and control algorithms. Finally, the simulation results are also presented and analyzed to demonstrate how the proposed simulation platform can facilitate the design and development of magnetic robotic systems.

### A. Bronchoscopy

Bronchoscopy is a minimally invasive procedure in which a flexible bronchoscope is inserted through the nose or mouth and advanced into the lungs to perform diagnosis, biopsy, or interventional tasks. Traditional flexible bronchoscopes are mainly steered by tendon-driven mechanisms, which increase the scope size and limit maneuverability within the narrow airways of the deep lung segments. Magnetic actuation, as a contactless method, can dexterously manipulate the distal tip of the bronchoscope and enable it to traverse otherwise inaccessible airways such as segmental bronchi.

In the simulation case of magnetically navigated bronchoscopy, a rigid and stationary single-layer geometric model is used as the simulation environment, which includes the main bronchus, lobar bronchi, and segmental bronchi. A permanent magnet is embedded in the distal tip of the flexible bronchoscope robot, with its magnetic moment aligned with the robot tip. The embedded magnet has a magnetic moment of $0.03$ A·m$^2$. The flexible bronchoscope is modeled as a continuum robot with an outer diameter of 2 mm. The parameters of the bronchoscope robots in simulation are listed in Table IV. An external permanent magnet is controlled by a 6-joint robot manipulator to exert magnetic torque on the robot for tip bending. The magnetic moment of the actuating magnet is initially set to be 350 A·m$^2$.

TABLE IV
Basic Robot Parameters in Three Use Cases

| Task | Total length (mm) | Outer Diameter (mm) | Inner Diameter (mm) | Young's Modulus (MPa) | Magnetic moment (A·m$^2$) |
|---|---|---|---|---|---|
| Case I | 600 | 2.0 | 1.5 | 100 | 0.030 |
| Case II | 1000 | 1.6 | 0.9 | 160 | 0.030 |
| Task | Mass (g) | Moment of Inertia (mg·m$^2$) | Outer Diameter (mm) | Length (mm) | Magnetic moment (A·m$^2$) |
| Case III | 8.60 | (0.6,0.6,0.6) | 12 | 25 | 0.13 |

The friction coefficient between the bronchoscope robot and the bronchus is set to be 0.015. The distance of collision detection is set to be 2 mm. The robot starts from the upper trachea and navigates through the airways until it reaches the target point at the lobar bronchi. In this case, the bronchoscope is controlled in a teleoperation mode, i.e., the desired motion of the robot including advancement and bidirectional bending of the robot tip is controlled by the user through an input device. Specifically, robot advancement is commanded by a keyboard. The desired bending direction of the robot is intuitively controlled by setting the direction of the magnetic field using a haptic device (Geomagic Touch, *3D Systems*), as shown in

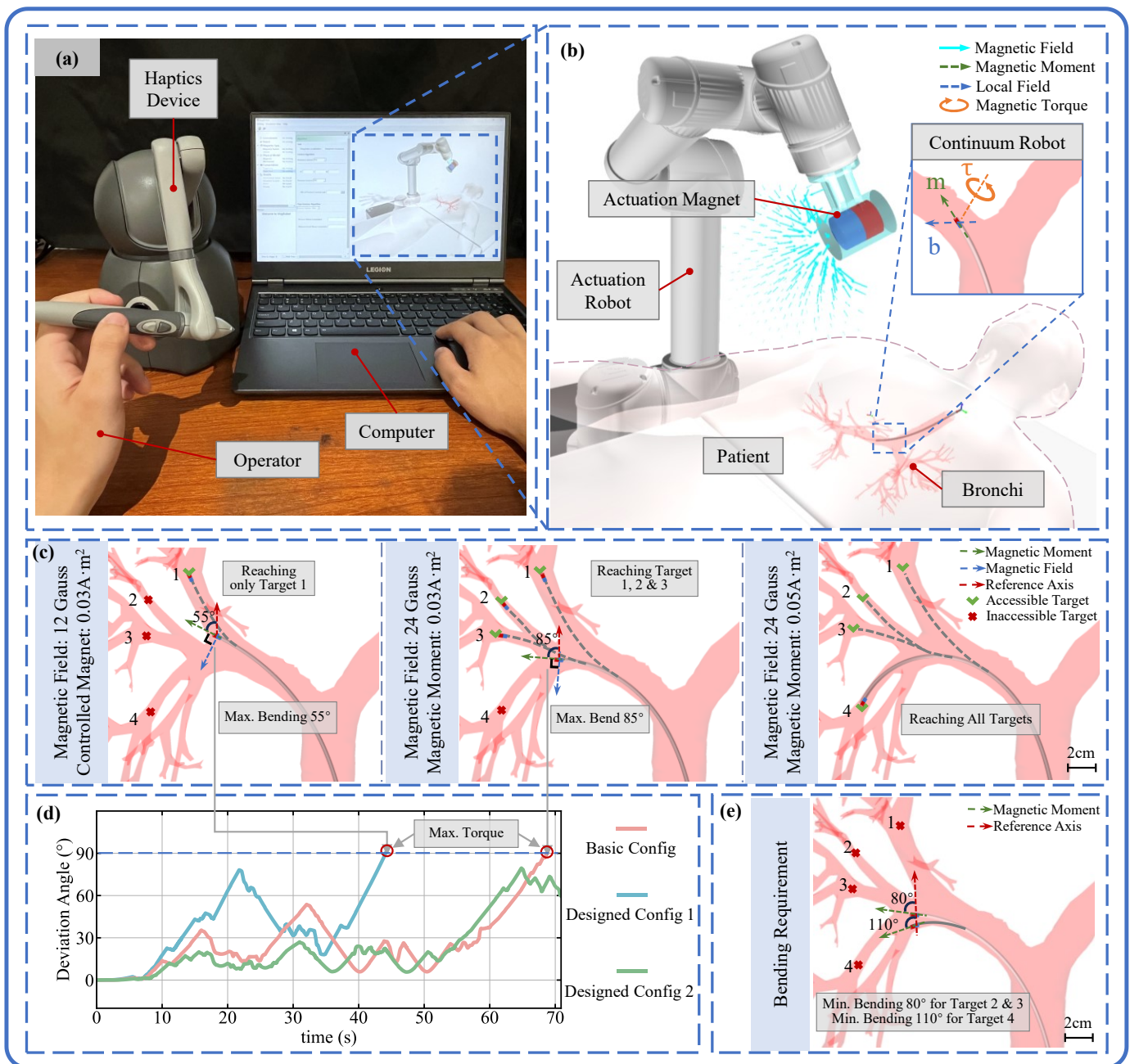

Fig. 8. Use case I: Magnetic actuation of a continuum robot in bronchoscopy. (a) Teleoperation setup for the user. (b) Simulation scene. (c) Simulation results of the basic and designed configurations. (d) Deviation angle of the robot tip from the direction of the actuating magnetic field. (e) Bending angle requirement to access targets.

TABLE V
Setup Parameters of Simulation in Three Use Cases

| Task | Environment | Robot | Magnetic Actuation Source | Trajectory | Motion Control Algorithm | Localization Algorithm | Friction Coefficient | Control Frequency(Hz) |
|---|---|---|---|---|---|---|---|---|
| Case I | Bronchi (rigid) | Continuum Robot | Manipulator-controlled PM | Teleoperation | Orientation: Human-in-the-loop Lateral Position: No control | — | 0.015 | 33 |
| Case II | Aortic Arch (soft) | Continuum Robot | Stationary EM | Extracted centerline | Orientation: PID Lateral Position: PID* | EKF | 0.01 | 200 |
| Case III | Stomach (soft) | Capsule Robot | Manipulator-controlled PM | Interpolation between waypoints | Orientation: Open-loop control Position: PID | — | 0.1 | 50 |

* Lateral position control is only implemented in the improved algorithm in Case II.

Fig.8 (a). The simulation is run at an update rate of 33 Hz. The setup parameters of the magnetic navigation system for the bronchoscope robot are given in Table V. The intensity of magnetic field is initially set to be 12 Gauss.

As shown in Fig. 8 (c), owing to insufficient actuation field, the bronchoscope robot can only reach Target 1. When attempting to bend into the branch leading to Target 2, the robot tip can only bend 55° at most relative to the vertical reference direction, resulting in the failure of accessing Target 2, 3, and 4. Fig. 8 (e) shows that reaching Targets 2 and 3 requires at least 80° deflection, and reaching Target 4 requires at least 110° deflection. The maximum bending angle (55°) is achieved when the deviation angle (the angle between the local field and the robot tip direction) reaches 90° to give the maximum magnetic torque, as shown in Fig. 8 (d). In the basic case with a magnetic field intensity of 12 Gauss, the bending angle cannot be further increased by controlling the direction of the actuation field. Therefore, the actuation field intensity at the robot tip is doubled to 24 Gauss by increasing the magnetic moment of the actuation magnet to 700 A·m² in the designed configuration 1. As shown in the middle plot of Fig. 8 (c), the robot can navigate to Target 1, 2, and 3 by intensifying the actuation field while the robot still fails to access Target 4 with an insufficient maximum bending angle of 85°. It is not desired to further increase the magnetic intensity of the actuation magnet due to the limit of size and weight of the actuation magnet. As a result, in the designed configuration 2, the magnetic moment of the embedded magnet on the robot tip is increased to 0.05 A·m² to obtain sufficient magnetic bending torque. With this final design of the magnetic actuation system, the robot can navigate to all four targets, as shown in the right plot of Fig. 8 (c).

This simulation of magnetic bronchoscopy demonstrates that the proposed simulator can successfully simulate magnetic navigation of a continuum robot within a rigid and static anatomy using a teleoperation control mode. It is also shown that the platform can be readily interfaced with input devices such as keyboards and haptic devices. Its post-processing module can provide abundant statistics and plots (e.g., the deviation angle in Fig. 8 (d)) to guide the design of magnetic navigation systems.

*B. Endovascular Intervention*

Endovascular interventions are minimally invasive procedures using a flexible catheter passing through arteries and veins to diagnose and treat vascular diseases. Compared with traditional open surgeries, endovascular procedures cause less incisions, shorter hospitalization, and lower risk of complications. In robot-assisted endovascular interventions, the catheter is driven by a delivery mechanism autonomously or in a tele-operation mode. Magnetic navigation system can improve the dexterity of the catheter robot by steering the tip segment and thus make the catheter pass the vascular system more fluently, efficiently, and safely.

This case simulates a catheter robot to navigate through the aortic arch under simultaneous magnetic actuation and localization. The initial configuration is shown in Fig. 9 (a). A deformable geometric model of the aortic arch vessel is employed as the simulation environment. The flexible catheter is modeled as a continuum robot, whose parameters are listed in Table IV. Whenever the robot contacts the aortic vessel, the deformation of both the catheter robot and the vessel as well as their interaction force are computed, assisting the user in evaluating procedural safety. A permanent magnet with a magnetic moment of 0.04 A·m² is embedded in the catheter tip. Instead of a manipulator-held permanent magnet, a stationary array of three electromagnetic coils is used as the actuation source, whose currents are controlled to generate the necessary magnetic field and gradient for catheter steering. An array of magnetic sensors measures the field produced by the embedded magnet for magnetic localization. The sensor noise level is set as 13 nT. Although the actuation coils also contribute to the sensor readings, their fields can be computed from the known coil currents and removed in the localization algorithm.

The aortic centerline is extracted by the trajectory generator module described in Section II B. The catheter robot is controlled automatically to follow this centerline through the aortic arch at a constant advancement velocity of 35 mm/s. In the initial simulation employing the three-coil configuration shown in Fig. 9 (a), magnetic localization of the catheter is performed using an extended Kalman filter, where 5-DoF pose of the catheter tip is estimated from magnetic measurements. The magnetic localization information serves as the feedback for magnetic control of the catheter robot. The catheter motion is controlled by the default control algorithm given in Section III C. The orientation of the catheter tip is controlled by a magnetic steering torque using a PID controller to follow the tangent direction of the centerline while the magnetic force is not controlled. The friction coefficient between the catheter and the aortic wall is 0.01. The distance of collision detection is 1mm. The simulation runs at an update rate of 200 Hz. The setup parameters of the magnetic navigation system for the catheter robot are listed in Table V.

Fig. 9 (b) shows the simulation results of magnetic navigation of the catheter robot using the three-coil actuation system without controlling magnetic force. It is shown that the catheter can approximately navigate through the aortic arc. The heading of the catheter aligns with the centerline well in the straight section of the blood vessel. As it turns around the arc, the heading deviates from the direction of centerline because the maximum bending torque is not sufficient to align the catheter tip. As shown in Fig. 9 (c), the average orientation control error of 12.46° while the orientation tracking error is 0.15° due to the use of high-precision sensors. However, uncontrolled magnetic force can deflect the tip from the centerline, causing large position control errors up to 9.62 mm and frequent collisions with the vessel wall that might damage the blood vessel during the procedure. To solve this issue, closed-loop lateral position control is added to the original control algorithm, which uses magnetic force to push the catheter tip toward the centerline path. Because three coils are not sufficient to independently control the 5-DoF magnetic force and torque of a permanent magnet, lateral position control of the catheter robot is infeasible using this three-coil actuation system. Therefore, the number of actuation coils needs to be increased. In the designed configuration, a new six-coil configuration of electromagnetic actuation system is designed with six coils uniformly distributed and tilted 45° upward, as shown in Fig. 9 (d). Fig. 9 (e) and (f) show the navigation results with simultaneous orientation and lateral position control. It is demonstrated that the catheter tip can precisely follow the centerline with position and orientation control errors of 2.21 mm and 11.25°. The position and orientation tracking errors are 0.22 mm and 0.17°. Fig. 9 (e) also shows that there is no collision between the catheter and the blood vessel throughout the procedure.

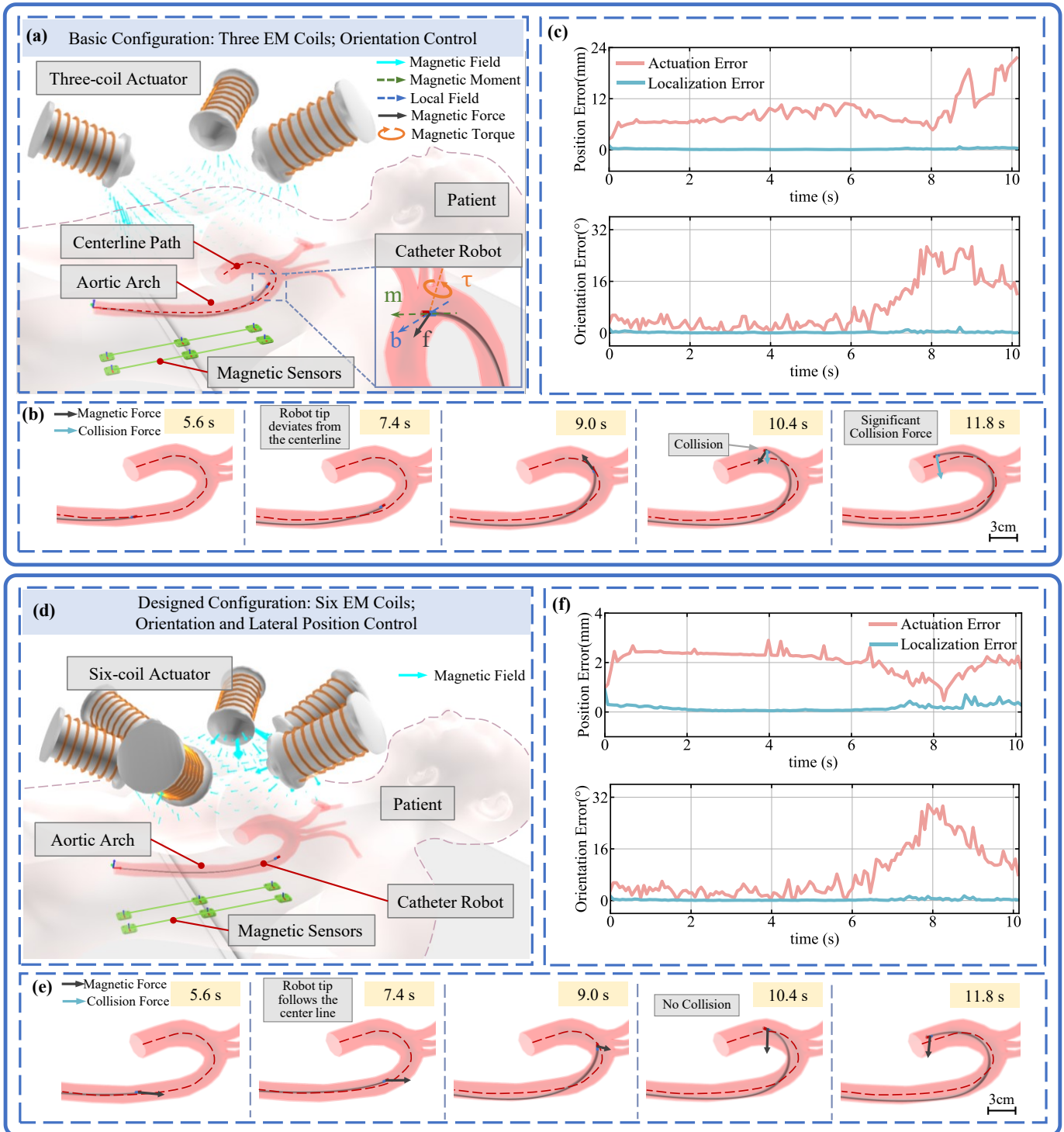

Fig. 9. Use case II: Simultaneous magnetic actuation and tracking of a catheter robot in cardiovascular intervention. (a) Simulation scene with the three-coil magnetic actuation system. (b) Temporal sequence of magnetic actuation of the catheter with only orientation control. (c) Magnetic actuation and tracking errors in the case with only orientation control. (d) Simulation scene with the eight-coil magnetic actuation system. (e) Temporal sequence of magnetic actuation of the catheter with both orientation and lateral position control. (f) Magnetic actuation and tracking error in the case with both orientation and lateral position control of the catheter tip.

This simulation case demonstrates that the proposed platform can simulate magnetic navigation of a continuum robot in a deformable environment under autonomous control mode. In addition, it is also demonstrated that the platform can simultaneously simulate the magnetic localization subsystem and the magnetic actuation subsystem. Finally, this case validates that the proposed platform can be readily used to analyze, design and optimize both the algorithm and hardware of the magnetic navigation system for better navigation performance.

*C. Capsule Endoscopy*

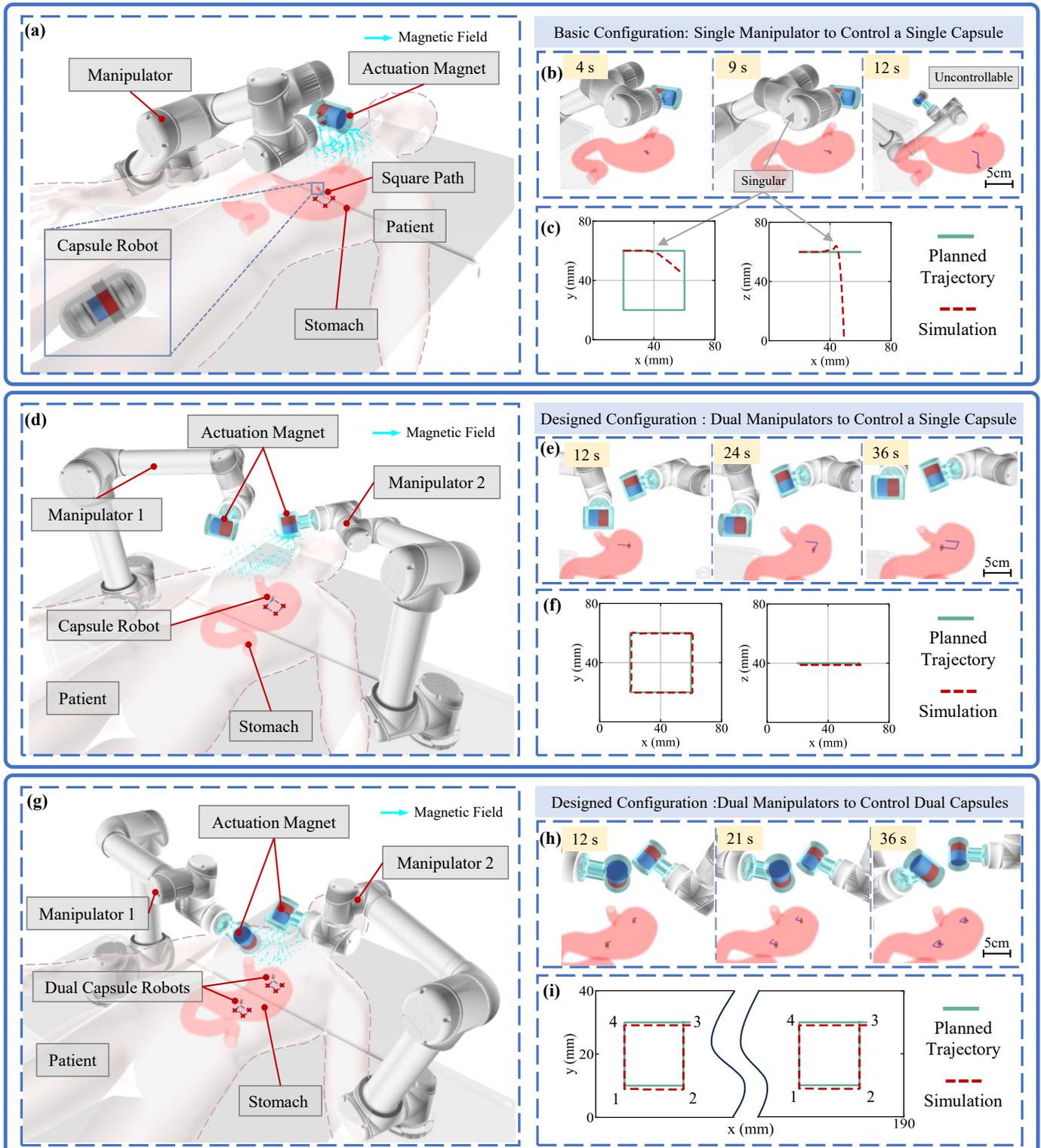

Fig. 10. Use case III: Magnetic actuation of a capsule robot in gastrointestinal endoscopy. (a) Simulation scene with a single capsule robot controlled by a single manipulator. (b) Temporal sequence of magnetic control of a single capsule robot using a single manipulator. (c) Trajectories of the single capsule robot controlled by a single manipulator. (d) Simulation scene with a single capsule robot controlled by dual manipulators. (e) Temporal sequence of magnetic control of a single capsule robot using dual manipulators. (f) Trajectories of the single capsule robot controlled by dual manipulators. (g) Simulation scene with dual capsule robots controlled dual manipulators. (h) Temporal sequence of magnetic control of dual capsule robots using dual manipulators. (i) Trajectories of dual capsule robots controlled by dual manipulators.

Capsule endoscopes are small-scale untethered devices with an embedded camera for inspecting the gastrointestinal tract for diagnostic purposes. Compared with traditional tethered endoscopy, capsule endoscopy can significantly reduce patient discomfort and access hard-to-reach lesions with higher precision. Magnetic actuation is an ideal remote actuation method for active locomotion and dexterous manipulation of capsule endoscopes.

This case simulates magnetic navigation of a capsule endoscope in the gastrointestinal tract. The initial configuration of magnetic actuation system is shown in Fig. 10 (a), which uses a single manipulator-controlled permanent magnet to control the capsule robot. A deformable geometric model of the stomach is employed as the simulation environment and the interaction between the endoscope and the anatomy can be simulated. The capsule endoscope is modeled as a rigid-body robot. A permanent magnet with a magnetic moment of 0.13 A·m² is embedded in the capsule endoscope for magnetic actuation. The parameters of the capsule robot are listed in Table IV. The actuation magnet has a magnet moment of 20 A·m² and is controlled by a 6-DOF robotic arm.

The friction coefficient between the gastrointestinal tract and the capsule robot is set as 0.1; the distance of collision detection is set to be 1 mm. A square trajectory is generated by interpolating user-defined waypoints, as shown in Fig. 10 (c). The robot is controlled to follow the square trajectory with a fixed orientation. The 3-DOF position of the capsule robot is controlled using a PID control law while its 2-DOF orientation is controlled in an open-loop manner, which sets the magnetic field direction at the capsule location as the desired orientation and thus align the heading of the capsule robot with the desired orientation. The simulation runs at 50 Hz. The setup parameters of the magnetic navigation system for the capsule robot are listed in Table V.

As shown in Fig. 10 (b), the single-magnet actuator controls the capsule robot well initially. However, the motion control diverges soon because the robotic arm reaches a singular pose, as shown in Fig. 10 (c). This singularity problem occurs due to the limited actuation degree-of-freedom of the single-magnet actuator. To increase the degree-of-freedom and redundancy of the magnetic actuation system, a new configuration of magnetic actuator is designed with two actuation magnets, each of which is controlled by a 6-DOF manipulator, as shown in Fig. 10 (d). The dual-magnet actuator can enlarge the workspace and reduce the occurrence of robot singularities. Fig. 10 (e) and (f) show the navigation results using the dual-magnet actuator. The capsule robot can accurately follow the entire square trajectory with an RMS position error of 1.42 mm. The dual-magnet actuator can also enable independent control of two capsule robots, which can collaborate to perform complicated tasks of diagnosis and treatment. As shown in Fig. 10 (g), another capsule robot can be readily added to the simulation scene. Fig. 10 (h) and (i) show that two capsule robots are controlled by the dual-magnet actuator to move along square trajectories independently, with RMS position control errors of 0.35 mm and 0.48 mm, respectively.

This simulation case demonstrates that the proposed platform can simulate magnetic navigation of rigid-body robots in a deformable anatomy using a hybrid control strategy (closed-loop position control and open-loop orientation control). This case also validates that the proposed platform can simulate complicated magnetic navigation systems involving multiple actuation magnets and multiple magnetic robots. The user can readily design novel magnetic navigation systems using the building blocks provided in the platform.

## VII. Conclusion

In this paper, a universal simulation platform, MagRobot, is proposed to facilitate research and design of magnetically navigated robots, especially for minimally invasive medicine. This work marks the first simulation platform that can simulate both magnetic actuation and magnetic tracking of rigid and soft magnetic robots in deformable anatomies for a variety of medical applications. In the simulator, the user can follow a standard workflow to efficiently design, visualize, and analyze magnetic navigation systems for medical robots through a user-friendly graphical interface. The simulator provides an open development environment, where the user can load third-party or customized models of environments (anatomies), design both hardware and algorithms of magnetic navigation systems, and export simulation data for cross-platform use. To extend the functionality and applications of the simulator and foster collaboration within the research community, we have made the simulator fully open-source.

For the simulator, we proposed a three-stage workflow of simulation, i.e., pre-processing, computation, and post-processing, and designed five key modules in the computation engine including trajectory generation, magnetic actuation, magnetic tracking, robot-environment interaction, and visual monitoring. Then we introduced the underlying physical models and default pose tracking/motion control algorithms of magnetic navigation systems. We validated the fidelity of the simulator using both phantom and *ex vivo* experiments of magnetic navigation of continuum and capsule robots with diverse magnetic actuation sources. It is shown that the simulated trajectories agree well with the experimental results. Finally, we implemented three use cases using the simulator, i.e., bronchoscopy, endovascular intervention, and gastrointestinal endoscopy, to demonstrate the functionality of the simulator. It is shown that the magnetic navigation systems can be efficiently configured for tracking and actuation of continuum and rigid robots with high fidelity, and the configuration parameters of magnetic navigation systems can be flexibly designed and optimized for better performance.

Although the proposed simulator has been shown to achieve reliable simulation of a wide range of magnetic navigation tasks for diverse magnetic robots, it still has some limitations. First, the current simulator does not support magnetic navigation of micro- and nano-robots, which are widely used for targeted drug delivery and micro-manipulation of biological cells. These robots operate under different physical models compared to macro-scale robots, necessitating specialized simulation capabilities. Second, the simulator does not account for dynamic fluid flow (e.g., blood flow) and its impact on robot behavior. In the future, we plan to integrate these missing features into the simulator and continue to collaborate with the research community to expand its functionalities and applications. Looking ahead, we will also connect the simulator with haptic devices and virtual reality technology to make a virtual training platform for health professionals to learn and practice magnetic navigation of medical robots in minimally invasive medicine.